\documentclass[10pt,twocolumn,letterpaper]{article}

\usepackage{cvpr}
\usepackage{times}
\usepackage{epsfig}
\usepackage{graphicx}
\usepackage{amsmath}
\usepackage{amssymb}
\usepackage{graphicx}
\usepackage{amsfonts}
\usepackage{physics}
\usepackage{booktabs}
\usepackage{siunitx}
\usepackage{graphicx,subfigure}
\usepackage{amssymb}
\usepackage{mathtools}
\usepackage{enumitem}
\usepackage{amsthm}
\usepackage[ruled,linesnumbered,noend]{algorithm2e}
\usepackage[noend]{algpseudocode}
\usepackage{wrapfig}
\usepackage{lettrine}
\usepackage{mdframed}
\usepackage{bm}
\usepackage{afterpage}

\newtheorem{definition}{Definition}
\newtheorem{theorem}{Theorem}

\newtheorem*{remark}{Remark}

\newcommand{\realNumber}{\mathbb{R}}
\newcommand{\naturalNumber}{\mathbb{N}}
\newcommand{\integerNumber}{\mathbb{Z}}

\newcommand{\commentout}[1]{}
\newcommand{\correction}[1]{{#1}}

\usepackage{hyperref}

\cvprfinalcopy 

\def\cvprPaperID{9636} 

\begin{document}

\title{Robustness Guarantees for Deep Neural Networks on Videos}

\author{Min Wu and Marta Kwiatkowska\\
Department of Computer Science, University of Oxford, UK\\
{\tt\small \{min.wu, marta.kwiatkowska@cs.ox.ac.uk\}}
}

\maketitle

\begin{abstract}
    The widespread adoption of deep learning models places demands on their robustness. In this paper, we consider the robustness of deep neural networks on \emph{videos}, which comprise both the spatial features of individual frames extracted by a convolutional neural network and the temporal dynamics between adjacent frames captured by a recurrent neural network. To measure robustness, we study the \emph{maximum safe radius} problem, which computes the minimum distance from the \emph{optical flow} \correction{sequence} obtained from a given input to that of an adversarial example in the \correction{neighbourhood of the input}. 
    We demonstrate that, under the assumption of Lipschitz continuity, the problem can be approximated using finite optimisation via discretising the optical flow space, and the approximation has provable guarantees. We then show that the finite optimisation problem can be solved by utilising \emph{a two-player turn-based game} in a cooperative setting, where the first player selects the optical flows and the second player determines the dimensions to be manipulated in the chosen flow. We employ an \emph{anytime} approach to solve the game, in the sense of approximating the value of the game by monotonically improving its upper and lower bounds. We exploit a gradient-based search algorithm to compute the upper bounds, and the admissible A* algorithm to update the lower bounds. Finally, we evaluate our framework on the UCF101 video dataset. 
\end{abstract}


\newcommand{\playerone}{\mathsf{Player}~\mathtt{I}}
\newcommand{\playertwo}{\mathsf{Player}~\mathtt{II}}

\section{Introduction}

Deep neural networks (DNNs) have been developed for a variety of tasks, including self-driving cars, malicious software classification, and abnormal network activity detection. While the accuracy of neural networks has significantly improved, matching human cognitive perception, they are susceptible to adversarial examples. An \emph{adversarial example} is an input which, whilst initially classified correctly, is misclassified with a slight, often imperceptible, perturbation. 
\emph{Robustness} of neural networks has been an active topic of investigation, and a number of approaches have been proposed (see Related Work below.)
However, most existing works focus on robustness of neural networks on image classification problems, where convolutional neural networks (CNNs) are sufficient. One assumption that CNNs rely on is that inputs are independent of each other, and they are unable to accept a \emph{sequence} of input data when the final output is dependent on intermediate outputs. In reality, though, tasks often contain sequential data as inputs, for instance, in machine translation~\cite{sutskever2014sequence}, speech/handwriting recognition~\cite{graves2013speech,graves2005framewise,fernandez2007application}, and protein homology detection~\cite{hochreiter2007fast}. 
To this end, \emph{recurrent neural networks (RNNs)} come into play. For RNNs, the connections between neurons form a directed graph along a temporal sequence, which captures temporal dynamic behaviours. Unlike CNNs, RNNs can use the internal state (memory) to process sequential inputs.

In this work, we study robustness guarantees for neural networks, including CNNs and RNNs, on \emph{videos}. 
Video classification is challenging because it comprises both the spatial features on each individual frames, which can be extracted by CNNs, as well as the temporal dynamics between neighbouring frames, which can be captured by RNNs.
Specifically, we develop a methodology for evaluating robustness of videos based on the notion of \emph{maximum safe radius} \correction{(or dually, minimum adversarial distortion~\cite{WZCYSGHD2018})}, which captures the maximum allowed magnitude of a perturbation. Our method is based on a game-based approach of \cite{wu2019game} and provides guarantees against perturbations up to a given magnitude. 
We compute upper and lower bounds of the maximum safe radius, and demonstrate their convergence \correction{trend} on the UCF101 video dataset.

\newcommand{\Reluplex}{\mathsf{Reluplex}}
\newcommand{\Planet}{\mathsf{Planet}}
\newcommand{\Sherlock}{\mathsf{Sherlock}}
\newcommand{\DLV}{\mathsf{DLV}}
\newcommand{\DeepGame}{\mathsf{DeepGame}}
\newcommand{\AI}{\mathsf{AI^2}}
\newcommand{\FastLin}{\mathsf{FastLin}}
\newcommand{\FastLip}{\mathsf{FastLip}}
\newcommand{\Crown}{\mathsf{Crown}}
\newcommand{\POPQORN}{\mathsf{POPQORN}}

%
\noindent
\textbf{Related work} \quad
\correction{
Robustness of neural networks has been mainly studied in the context of image classification, and, to the best of our knowledge, very few works address robustness guarantees for videos. We review only approaches most relevant to our approach.
Early methods evaluate the robustness of networks by checking whether they are vulnerable to \emph{adversarial attacks}, which attempt to craft a perturbation imposed on the input so that it is misclassified by a well-trained network.
In the computer vision and security communities, various attack techniques have been developed, such as the limited-memory Broyden-Fletcher-Goldfarb-Shanno algorithm~\cite{szegedy2014intriguing}, the fast gradient sign method~\cite{goodfellow2015explaining} and its extended Jacobian saliency map based attack~\cite{papernot2016limitations}, as well as the optimisation-based C\&W attack~\cite{carlini2017towards}.
Nevertheless, adversarial attacks cannot provide guarantees.
To this end, various \emph{verification}  approaches for neural networks emerged in the formal methods community.
These include constraint solving tools $\Reluplex$~\cite{katz2017reluplex} and $\Planet$~\cite{ehlers2017formal}; mixed integer linear programming method $\Sherlock$~\cite{dutta2018output}; the $\AI$ approach~\cite{gehr2018ai2} involving abstract interpretation; algorithms utilising linear approximation such as $\FastLin$/$\FastLip$~\cite{weng2018towards} and $\Crown$~\cite{zhang2018efficient}; and also search-based $\DLV$~\cite{huang2017safety} and game-based tool $\DeepGame$~\cite{wu2019game}.

The above approaches evaluate the robustness of feedforward networks, but we are interested in providing guarantees for both convolutional and recurrent layers with time-series inputs.
Although there are some adversarial attacks against RNNs~\cite{papernot2016crafting} and videos~\cite{inkawhich2018adversarial,wei2018sparse}, verification approaches are rare. For example, $\POPQORN$~\cite{ko2019POPQORN} employs the linear approximation method, bounding the network output by two linear functions in terms of input; and Wang et al.~\cite{wang2018verification} extract deterministic finite automata from RNNs as the oracle and use them to evaluate adversarial accuracy.
Our approach instead draws on $\DeepGame$~\cite{wu2019game}, where a game-based verification framework is proposed for computing the maximum safe radius for feedforward networks through discretising the input space via Lipschitz continuity~\cite{hein2017formal}, but in this work we are able to handle CNN + RNN architectures and video inputs.
}

\newcommand{\pixel}{\mathcal{P}}
\newcommand{\distance}[2]{\norm{#1}_{#2}}
\newcommand{\ball}{\mathsf{B}}
\newcommand{\Frame}{\mathcal{F}}
\newcommand{\video}{\bm{v}}
\newcommand{\network}{\mathcal{N}}
\newcommand{\cnn}{\mathsf{C}}
\newcommand{\rnn}{\mathsf{R}}
\newcommand{\class}{c}
\newcommand{\classes}{C}
\newcommand{\domain}{\mathrm{D}}
\newcommand{\Lipschitz}{\mathsf{Lip}}

\setlength{\abovedisplayskip}{1pt}
\setlength{\belowdisplayskip}{0pt}

\section{Preliminaries}

\subsection{Deep neural networks}

Let $\network$ be a neural network with a set of classes $\classes$. Given an input $\video$ and a class $\class \in \classes$, we use $\network (\video, \class)$ to denote the confidence of $\network$ believing that $\video$ is in class $\class$. We work with the $\mathsf{Softmax}$ logit value of the last layer, but the methods can be adapted to the probability value after normalisation. Thus, $\network (\video) = \arg \max_{\class \in \classes} \network (\video, \class)$ is the class into which $\network$ classifies $\video$. Moreover, as $\network$ in this work can have convolutional and recurrent layers, we let $\network_\cnn$ denote the convolutional part and $\network_\rnn$ the recurrent part.  
Specifically, since the inputs we consider are videos, we let the input domain $\domain$ be $\realNumber^{l \times w \times h \times ch}$, where $l$ is the length of $\video$, i.e., the number of frames, and $w, h, ch$ are the width, height, and channels of each frame, respectively.

\subsection{Optical flow}

In order to capture the dynamic characteristics of the moving objects in a video, we utilise \emph{optical flow}~\cite{burton1978thinking,warren2013electronic}, which is a pattern of the apparent motion of the image objects between two consecutive frames caused by the movement of the objects or the camera.
\correction{Several methods exist in the computer vision community to compute flows, e.g., the Lucas-Kanade~\cite{lucas1981iterative} and Gunnar Farneb{\"a}ck~\cite{farneback2003two} algorithms.}

\begin{definition}[Optical Flow]
    Consider a pixel $\pixel(x,y,t)$ in a frame, where $x,y$ denote the horizontal and vertical positions respectively, and $t$ denotes time dimension. If after $\dd{t}$ time, the pixel moves by distance $(\dd{x},\dd{y})$ in the next frame, then 
        $\pixel(x,y,t) = \pixel(x+\dd{x}, y+\dd{y}, t+\dd{t})$
    holds. After taking Taylor series approximation, removing common terms, and dividing by $\dd{t}$, the \emph{Optical Flow Equation} is
        $f_x u + f_y v + f_t = 0, 
        \ \text{such that} \
        f_x = \frac{\partial f}{\partial x}, \ f_y = \frac{\partial f}{\partial y}, \
        u = \frac{\partial x}{\partial t}, \ v = \frac{\partial y}{\partial t},$
    where $f_x, f_y$ are the image gradients, $f_t$ is the gradient along time, and the motion $(u,v)$ is unknown.
\end{definition}

\subsection{Distance metrics and Lipschitz continuity}

In robustness evaluation, $L^p$ distance metrics are often used to measure the discrepancy between inputs, denoted as $\distance{\alpha-\alpha'}{p}$, where $p \in \{ 1,2,\infty \}$ indicates Manhattan ($L^1$), Euclidean ($L^2$), and Chebyshev ($L^\infty$) distances. Because our inputs are videos, i.e., sequences of frames, we will need a suitable metric. In this work, we work directly with $L^p$ metrics on \emph{optical flows}, as described in the next section.
Moreover, we consider neural networks that satisfy \emph{Lipschitz continuity}, and note that all networks with bounded inputs are Lipschitz continuous, such as the common fully-connected, convolutional, ReLU, and softmax layers. We denote by $\Lipschitz_\class$ the \emph{Lipschitz constant} for class $\class$.

\newcommand{\spatial}{\eta}
\newcommand{\flow}{\mathtt{p}}
\newcommand{\flowset}{\mathsf{P}}
\newcommand{\robust}{\textsc{Robust}}
\newcommand{\loss}{\mathtt{l}}
\newcommand{\manipulation}{\mathcal{M}}
\newcommand{\instruction}{\Theta}
\newcommand{\atomic}{\theta}
\newcommand{\magnitude}{\tau}
\newcommand{\MSR}{\mathtt{MSR}}
\newcommand{\FMSR}{\mathtt{FMSR}}
\newcommand{\dist}{\Tilde{d}}
\newcommand{\grid}{\Gamma}
\newcommand{\gridPoint}{\overline{g}}

\section{Robustness: formulation and approximation}
In this section, we formulate the robustness problem and provide an approximation with provable guarantees.

\subsection{Robustness and maximum safe radius}

In this work, we focus on \emph{pointwise robustness}, which is defined as the invariance of a network's classification over a small neighbourhood of a given input. Following this, the \emph{robustness} of a classification decision for a specific input can be understood as the non-existence of adversarial examples in the neighbourhood of the input. 
Here, we work with the norm ball as a neighbourhood of an input, that is, given an input $\video$, a distance metric $L^p$, and a distance $d$, $\ball(\video, L^p, d) = \{ \video' \mid \distance{\video-\video'}{p} \leq d \}$ is the set of inputs whose distance to $\video$ is no greater than $d$ based on the $L^p$ norm. 
Intuitively, the norm ball $\ball$ 
with centre $\video$ and radius $d$ limits perturbations to at most $d$ with respect to $L^p$. Then (pointwise) robustness is defined as follows.

\begin{definition}[Robustness] \label{dfn:robustness}
    Given a network $\network$, an input $\video$, a distance metric $L^p$, and a distance $d$, an \emph{adversarial example} $\video'$ is such that $\video' \in \ball(\video, L^p, d)$ and $\network(\video') \neq \network(\video)$. Define the \emph{robustness} of $\video$ by 
        $\robust(\network, \video, L^p, d) \models \nexists \ \video' \in \ball(\video, L^p, d) \ \text{s.t.} \ \network(\video') \neq \network(\video)$.
    If this holds, we say $\network$ is safe w.r.t. $\video$ within $d$ based on the $L^p$ norm.
\end{definition}

\begin{figure*}
    \begin{minipage}{0.3\linewidth}
        \centering
        \includegraphics[height=5cm]{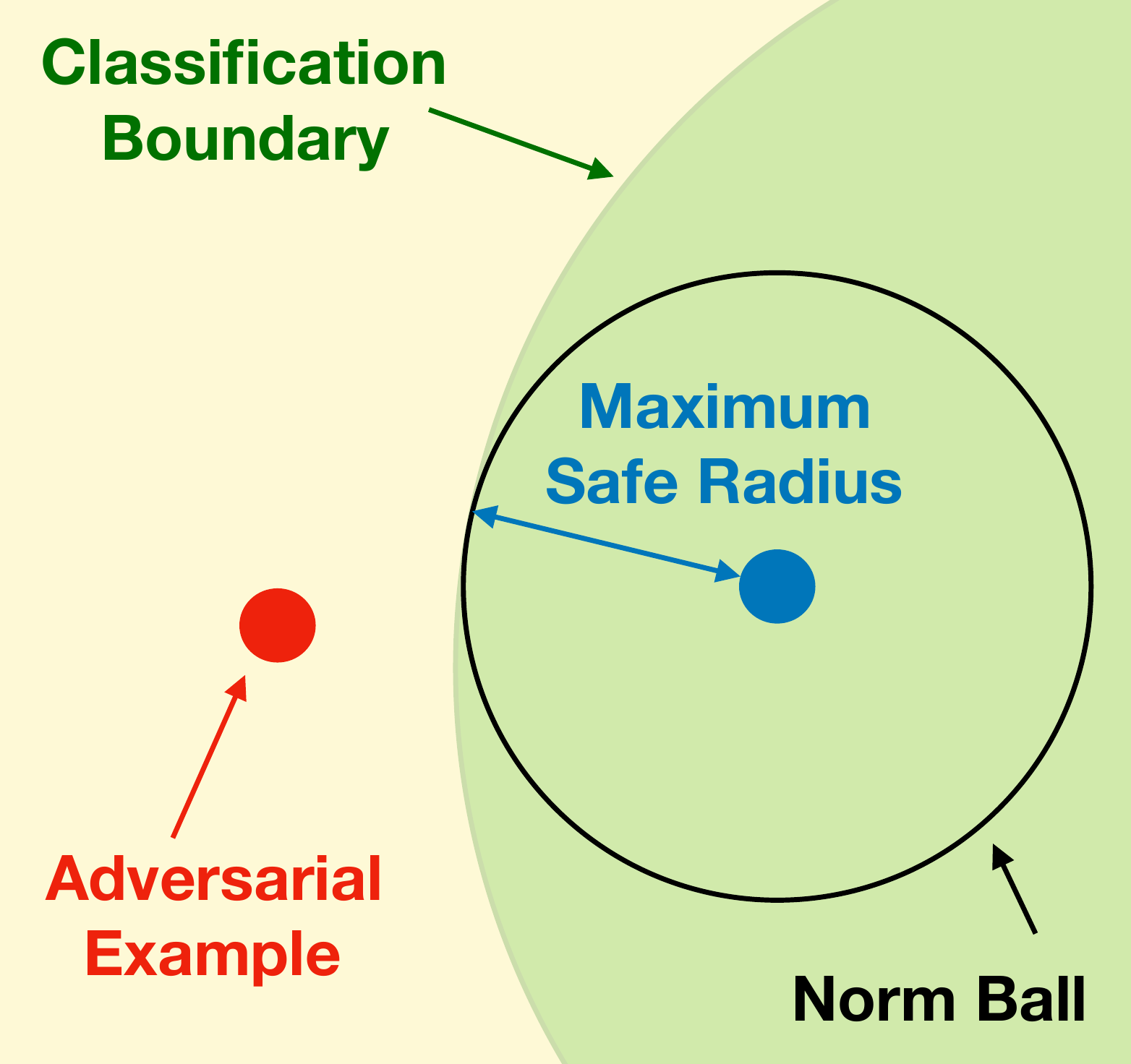}
        \caption{Maximum safe radius.}
        \label{fig:MSR}
    \end{minipage}
    \begin{minipage}{0.34\linewidth}
        \centering
        \includegraphics[height=5cm]{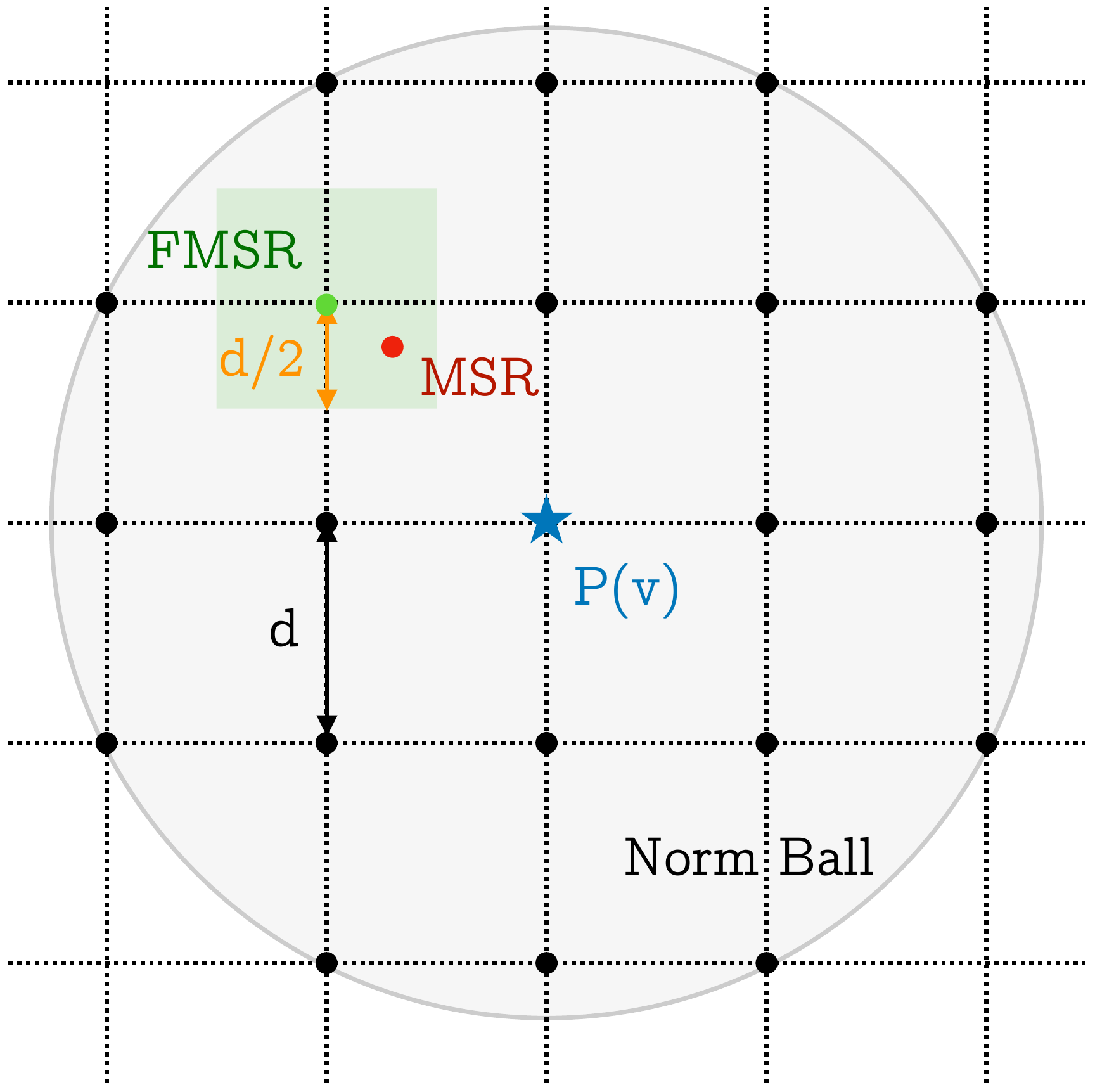}
        \caption{Error bounds.}
        \label{fig:error}
    \end{minipage}
    \begin{minipage}{0.36\linewidth}
        \centering
        \includegraphics[height=5cm]{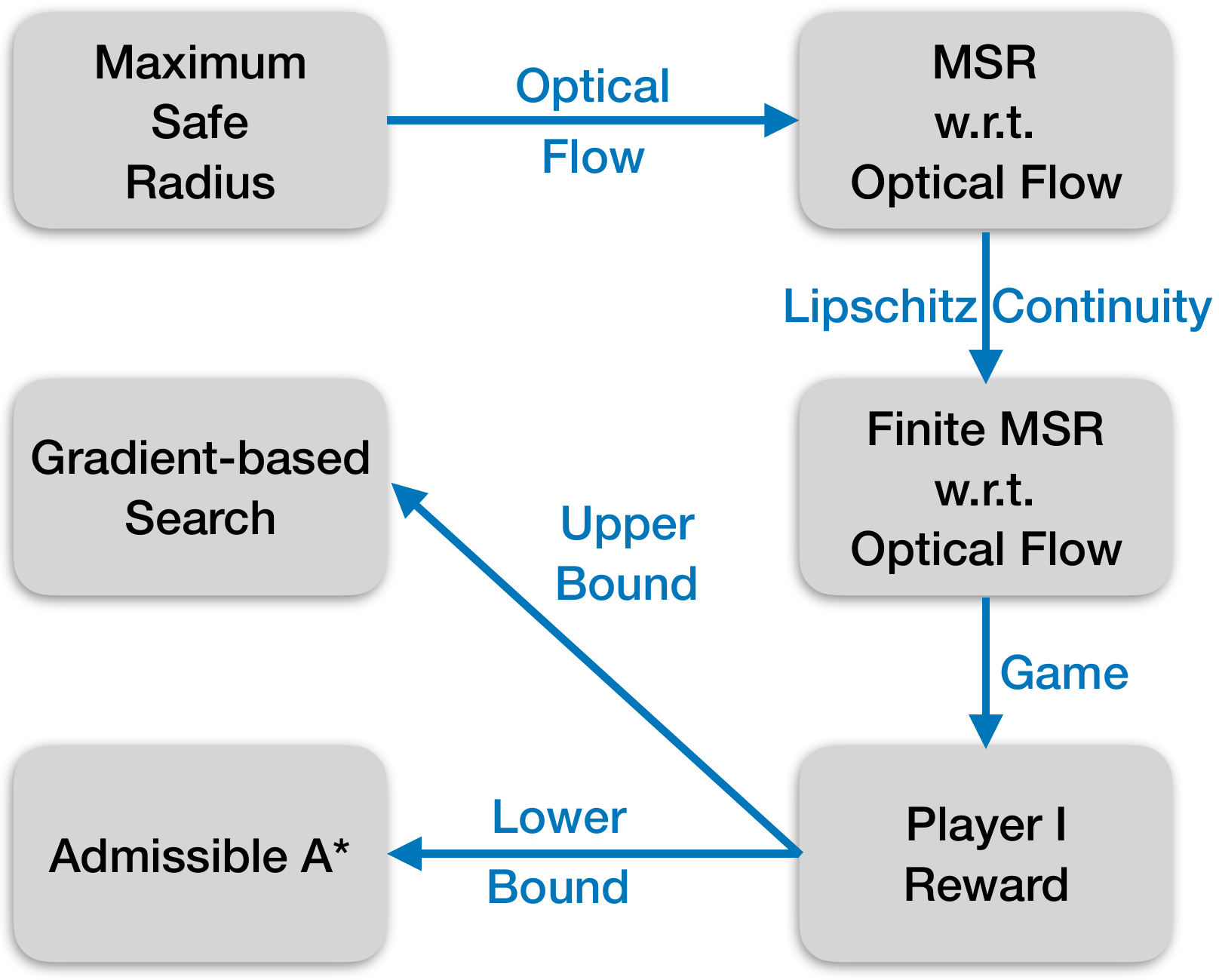}
        \caption{\correction{Robustness verification workflow.}}
        \label{fig:workflow}
    \end{minipage}
    \vspace{-1em}
\end{figure*}

While the above definition returns only $\mathtt{True}$ or $\mathtt{False}$, we take a step further to quantify the measurement of robustness. That is, we compute the distance to the original input in the sense that, if exceeding the distance, there definitely exists an adversarial example, whereas, within the distance, all the points are safe, see Figure~\ref{fig:MSR}.
We formally define this distance as the \emph{maximum safe radius} as follows.

\begin{definition}[Maximum Safe Radius] \label{dfn:MSR}
    Given a network $\network$, an input $\video$, a distance metric $L^p$, and a distance $d$, the \emph{maximum safe radius} $(\MSR)$ problem is to compute the minimum distance from input $\video$ to an adversarial example $\video'$, i.e., 
    \correction{
    \begin{multline}
        \MSR(\network, \video, L^p, d) = 
        \min_{\video' \in \domain} \{ \distance{\video - \video'}{p}
        \mid \video' \in \ball(\video, L^p, d) \\ \text{such that} \ \ \network(\video') \neq \network(\video) \}.
    \end{multline}
    If $\video'$ does not exist in $\ball$, we let $\MSR = d+\epsilon$, i.e., $d$ plus a small value $\epsilon$.}
\end{definition}

\subsection{Maximum safe radius w.r.t. optical flow}

In existing works that evaluate a network's robustness over images, it is common to manipulate each image at pixel- or channel-level, and then compute the distance between the perturbed and original inputs. However, as we deal with time-series inputs, i.e., \emph{videos}, instead of manipulating directly on each individual frame, we impose perturbation on each \emph{optical flow} that is extracted from every pair of adjacent frames, so that both spatial features on frames and temporal dynamics between frames can be captured. 
\begin{definition}[Optical Flow] \label{dfn:flow}
    Given an input video $\video$ with number $l$ of frames, i.e., $\video = \{ \Frame_1, \ldots, \Frame_t, \ldots, \Frame_l \}, t \in [1, l], t \in \naturalNumber^+ $, the \emph{optical flow} extraction function $f: \Frame_t, \Frame_{t+1} \mapsto \flow_t$ maps every two adjacent frames $\Frame_t, \Frame_{t+1}$ into an optical flow $\flow_t$. Then, for the sequential video $\video$, a \emph{\correction{sequence} of optical flows} can be extracted, i.e., $\flowset(\video) = \{ \flow_1, \ldots, \flow_t, \ldots, \flow_{l-1} \}, t \in [1, l-1], t \in \naturalNumber^+$.
\end{definition}
We remark that the distance between the flow sequences of two videos, denoted as $\distance{\flowset(\video) - \flowset(\video')}{p}$, can be measured similarly to that of two videos $\distance{\video-\video'}{p}$ by the $L^p$ norms in the standard way, as they are essentially tensors.

Then, to study the crafting of adversarial examples, we construct \emph{ manipulations} on the optical flow to obtain perturbed inputs. 
Note that if the input values are bounded, e.g., $[0,255]$ or $[0,1]$, then the perturbed inputs need to be restricted to be within the bounds.

\begin{definition}[Optical Flow Manipulation] \label{dfn:manipulation}
    Given an input $\video$ with optical flow \correction{sequence} $\flowset(\video)$, an instruction function $\instruction: \correction{\naturalNumber^+  \rightarrow \integerNumber}$, and a manipulation magnitude \correction{$\magnitude \in \realNumber$}, we define the \emph{optical flow manipulation} 
    \correction{
    $\manipulation_{\instruction,\magnitude} (\flow_t)(i) =$
    \begin{equation}
        \begin{cases}
            \flow_t[i] + \instruction(i) \cdot \magnitude,    & \quad \text{if } i \in [1, w \times h], i \in \naturalNumber^+ \\
            \flow_t[i],    & \quad \text{otherwise}
        \end{cases}
    \end{equation}
    where $w,h$ denote the width and height, i.e., $\instruction$ \correction{acts on} flow dimensions. In particular, when $\atomic: \naturalNumber^+ \rightarrow \{+1,-1\}$, we say the manipulation is \emph{atomic}, denoted as $\manipulation_{\atomic,\magnitude}$.}
\end{definition}
\correction{Instead of manipulating input $\video$ directly, we manipulate the optical flow using $\manipulation_{\instruction,\magnitude}(\flowset(\video))$, and then impose the manipulated flow on the original frames to obtain the corresponding perturbed input $\video'$ such that $\flowset(\video') = \manipulation_{\instruction,\magnitude}(\flowset(\video))$.}
To this end, we compute the distance from $\manipulation_{\instruction,\magnitude}(\flowset(\video))$ to $\flowset(\video)$ instead of that from $\video'$ to $\video$ because the former reflects both spatial and temporal manipulations. That is, we compute the \emph{maximum safe radius w.r.t. optical flow} $\MSR(\network,\flowset(\video),L^p,d)$
\correction{rather than $\MSR(\network,\video,L^p,d)$.}

\subsection{Approximation based on Lipschitz continuity}

Here, we utilise the fact that the networks studied in this work are \emph{Lipschitz continuous}, which allows us to discretise the neighbourhood space of an optical flow \correction{sequence and consider a finite number of points on a grid instead of infinitely many.} 
First, based on the definitions of optical flow and input manipulation, we transform the $\MSR$ problem into the following \emph{finite maximum safe radius} problem. 
\begin{definition}[Finite Maximum Safe Radius] \label{dfn:FMSR}
    Given an input $\video$ with the optical flow \correction{sequence} $\flowset(\video)$ and a manipulation operation $\manipulation_{\instruction,\magnitude}$, then the \emph{finite maximum safe radius ($\FMSR$) w.r.t. optical flow} is
    \begin{multline} \label{eqn:FMSR}
        \FMSR(\network, \flowset(\video), L^p, d, \magnitude) = 
        \min_{\flow_t \in \flowset(\video)} \min_{\atomic \in \instruction} \{
        \\
        \distance{\flowset(\video) - \correction{\manipulation_{\atomic,\magnitude}(\flowset(\video))}}{p}
        \mid \manipulation_{\atomic,\magnitude}(\flowset(\video)) \in
        \ball(\flowset(\video), L^p, d)
        \\
        \text{s.t.} \ \network(\video') \neq \network(\video)
        \correction{\ \text{and} \ \flowset(\video') = \manipulation_{\atomic,\magnitude}(\flowset(\video))
        \} .
        }
    \end{multline}
    \correction{If $\video'$ does not exist in $\ball$, we let $\FMSR = d+\epsilon$.}
\end{definition}

Intuitively, we aim to find a set of \correction{atomic} manipulations $\atomic \in \instruction$ to impose on a \correction{sequence} of optical flows $\flow_t \in \flowset(\video)$, such that the distance between the flow \correction{sequences} is minimal, and the corresponding perturbed input $\video'$ is an adversarial example. Considering that, within a norm ball $\ball(\flowset(\video),L^p,d)$, the set of manipulations is finite for a fixed magnitude $\magnitude$, the $\FMSR$ problem only needs to explore a finite number of the `grid' points. 
To achieve this, we let $\gridPoint$ be a $\magnitude$-grid point such that $\abs{\gridPoint - \flowset(\video)} = n \times \magnitude$, and $\grid(\flowset(\video), L^p, d)$ be the set of $\magnitude$-grid points whose corresponding optical flow \correction{sequences} are in $\ball$. Note that all the $\magnitude$-grid points are reachable from each other via manipulation. By selecting an appropriate $\magnitude$, we ensure that the optical flow space can be covered by small sub-spaces. That is, $\ball(\flowset(\video),L^p,d) \subseteq \bigcup_{\gridPoint \in \grid} \ball (\gridPoint, L^p, \frac{1}{2} \dist(L^p,\magnitude))$, where the grid width $\dist(L^p,\magnitude)$ is $|\domain|\magnitude$ for $L^1$, $\sqrt{|\domain|\magnitude^2}$ for $L^2$, and $\magnitude$ for $L^\infty$.
Now, we can use $\FMSR$ to estimate $\MSR$ within the error bounds, as shown in Figure~\ref{fig:error}.
\correction{We remark that approximation is only involved in the estimation of $\MSR$ from $\FMSR$, which does not compromise the guarantees provided by $\FMSR$.}

\begin{theorem}[Error Bounds] \label{thm:bounds}
    Given a manipulation magnitude $\magnitude$, the optical flow space can be discretised into a set of $\magnitude$-grid points, and $\MSR$ can be approximated as 
    \begin{multline} \label{eqn:bounds}
        \FMSR(\network, \flowset(\video), L^p, d, \magnitude) - \frac{1}{2} \dist(L^p, \magnitude)
        \\
        \leq \MSR(\network, \flowset(\video), L^p, d) \leq
        \FMSR(\network, \flowset(\video), L^p, d, \magnitude).
    \end{multline}
\end{theorem}

We next show how to determine $\magnitude$. Note that, in order to make sure each $\magnitude$-grid point $\gridPoint$ covers all the possible manipulation points in its neighbourhood, we compute the largest $\magnitude$, which can be obtained via \emph{Lipschitz continuity}. For a network $\network$ that is Lipschitz continuous at input $\video$, given Lipschitz constant $\Lipschitz_\class, \class \in \classes$, we have
\begin{equation} \label{eqn:Lipschitz}
    \dist'(L^p, \magnitude) \leq 
    \dfrac{\displaystyle \min_{\class \in \classes, \class \neq \network(\video)}
    \{ \network(\video, \network(\video)) - \network(\video, \class) \} }
    {\displaystyle \max_{\class \in \classes, \class \neq \network(\video)} 
    (\Lipschitz_{\network(\video)} + \Lipschitz_\class) }.
\end{equation}
\correction{The proof is given in Appendix~\ref{app:error}}.
%
Here we remark that, while $\dist'(L^p, \magnitude)$ is with respect to input $\video$ and $\dist(L^p, \magnitude)$ is with respect to the flow \correction{sequence} $\flowset(\video)$, the relation between them is dependent on the flow extraction method used. As this is not the main focus of this work, we do not expand on this topic.

\newcommand{\one}{\mathtt{I}}
\newcommand{\two}{\mathtt{II}}
\newcommand{\game}{\mathcal{G}}
\newcommand{\transition}{T}
\newcommand{\labelling}{L}
\newcommand{\rewards}{R}
\newcommand{\Path}{\rho}
\newcommand{\strategy}{\sigma}
\newcommand{\terminate}{tc}

\section{A game-based verification approach}

In this section, we show that the finite optimisation problem $\FMSR$ of Definition~\ref{dfn:FMSR} can be reduced to the computation of a player's reward when taking an optimal strategy in a game-based setting. 
The verification workflow is in Figure~\ref{fig:workflow}.
To this end, we adapt the game-based approach proposed in \cite{wu2019game} for robustness evaluation of CNNs on images.

\subsection{Problem solving as a two-player game}

We define a two-player turn-based game, where $\playerone$ chooses which optical flow to perturb, and $\playertwo$ then imposes atomic manipulations within the selected flow. 

\begin{definition}[Game] \label{dfn:game}
    Given \correction{a network $\network$}, an input $\video$ and its extracted optical flow \correction{sequence} $\flowset(\video)$, we let $\game(\network,\video,L^p,d) = (S \cup (S \times \flowset(\video)), s_0, \{ \transition_\one, \transition_\two \}, \labelling)$ be \emph{a game model}, where
    \vspace{-\topsep}
    \begin{itemize}[leftmargin=1em]
        \setlength{\parskip}{0pt}
        \setlength{\itemsep}{0pt plus 1pt}
        \item $S \cup (S \times \flowset(\video))$ denotes the \emph{set of game states}, in which $S$ is the set of $\playerone$'s states whereas $S \times \flowset(\video)$ is the set of $\playertwo$'s states. Each $s \in S$ corresponds to an optical flow \correction{sequence} $\flowset(s)$ in the norm ball $\ball(\flowset(\video), L^p, d)$.
        
        \item $s_0 \in S$ is the \emph{initial state} such that $\flowset(s_0)$ corresponds to the original optical flow \correction{sequence} $\flowset(\video)$.
        
        \item $\transition_\one: S \times \flowset(\video) \rightarrow S \times \flowset(\video)$ is $\playerone$'s \emph{transition relation} defined as
            $\transition_\one(s, \flow_t) = (s, \flow_t)$,
        and $\transition_\two: (S \times \flowset(\video)) \times \instruction \rightarrow S$ is $\playertwo$'s \emph{transition relation} defined as
            $\transition_\two((s,\flow_t), \atomic) = \manipulation_{\atomic, \magnitude}(\flow_t)$,
        where $\manipulation_{\atomic, \magnitude}$ is the atomic manipulation of Definition~\ref{dfn:manipulation}. 
        Intuitively, in a game state $s$, $\playerone$ selects a flow $\flow_t$ of $\flowset(s)$ and enters into a $\playertwo$'s state $(s,\flow_t)$, where $\playertwo$ then chooses an atomic manipulation $\manipulation_{\atomic,\magnitude}$ on $\flow_t$.
        
        \item $\labelling: S \cup (S \times \flowset(\video)) \rightarrow \classes$ is the \emph{labelling function} that assigns \correction{to each game state the corresponding class}. 
    \end{itemize}
\end{definition}

To compute $\FMSR$ of Definition~\ref{dfn:FMSR}, we let the game $\game$ be \emph{cooperative}.
When it proceeds, two players take turns -- $\playerone$ employs a strategy $\strategy_\one$ to select optical flow, then $\playertwo$ employs a strategy $\strategy_\two$ to determine atomic manipulations -- thus forming a path $\Path$, which is a sequence \correction{ of states and actions $s_0{\flow_{0}}s_1{\theta_1}s_2{\flow_{2}}s_3 \cdots$}.
Formally, we define the strategy of the game as follows.
    Let $Path^F_\one$ be a set of finite paths ending in $\playerone$'s state, and $Path^F_\two$ be a set of finite paths ending in $\playertwo$'s state, then we define a \emph{strategy profile} $\strategy=(\strategy_\one, \strategy_\two)$, such that $\strategy_\one : Path^F_\one \rightarrow \mathcal{D}(\flowset(\video))$ of $\playerone$ maps a finite path to a distribution over next actions, and similarly $\strategy_\two : Path^F_\two \rightarrow \mathcal{D}(\instruction)$ for $\playertwo$.

Intuitively, by imposing atomic manipulations in each round, the game searches for potential adversarial examples with increasing distance to the original optical flow. Given $\Path$, let \correction{$\video'_\Path$} denote the input corresponding to the last state of $\Path$, and $\flowset(\video'_\Path)$ denote its optical flow \correction{sequence}, we write the \emph{termination condition} 
    $\terminate(\Path) \equiv (\network(\video'_\Path) \neq \network(\video)) \vee (\distance{\flowset(\video'_\Path)-\flowset(\video)}{p} > d)$,
which means that the game is in a state whose corresponding input is either classified differently, or the associated optical flow \correction{sequence} is outside the norm ball. 
To quantify the distance accumulated along a path, we define a reward function as follows. Intuitively, the reward is the distance to the original optical flow if an adversarial example is found, and otherwise it is the weighted summation of the rewards of its children on the game tree.
\begin{definition}[Reward] \label{dfn:reward}
    Give a finite path $\Path$ and a strategy profile $\strategy = (\strategy_\one,\strategy_\two)$, we define a \emph{reward} function $\rewards(\strategy,\Path)$
    \begin{equation}
        =
        \begin{cases}
            \distance{\flowset(\video'_\Path)-\flowset(\video)}{p}, \\
            \quad \quad \quad \quad \quad \quad \quad
            \text{if \ } \terminate(\Path) \text{ and } \Path \in Path^F_\one \\
            \sum_{\flow_t \in \flowset(\video)} \strategy_\one(\Path)(\flow_t) \cdot \rewards(\strategy,\Path \transition_\one(last(\Path),\flow_t)), \\
            \quad \quad \quad \quad \quad \quad \quad
            \text{if \ } \neg \terminate(\Path) \text{ and } \Path \in Path^F_\one \\
            \sum_{\atomic \in \instruction} \strategy_\two(\Path)(\atomic) \cdot \rewards(\strategy,\Path \transition_\two(last(\Path),\atomic)), \\
            \quad \quad \quad \quad \quad \quad \quad
            \text{if \ } \Path \in Path^F_\two
        \end{cases}
    \end{equation}
    where $\strategy_\one(\Path)(\flow_t)$ is the probability of $\playerone$ choosing flow $\flow_t$ along $\Path$, and $\strategy_\two(\Path)(\atomic)$ is that of $\playertwo$ determining atomic manipulation $\manipulation_{\atomic,\magnitude}$ along $\Path$. Also, $\Path \transition_\one(last(\Path),\flow_t)$ and $\Path \transition_\two(last(\Path),\atomic)$ are the resulting paths of $\playerone$, $\playertwo$ applying $\strategy_\one$, $\strategy_\two$, respectively, i.e., extending $\Path$ by adding a new state after transition.
\end{definition}

\subsection{Robustness guarantees}

We now confirm that the game can return the optical value of the reward function as the solution to the $\FMSR$ problem.
Proof of Theorem~\ref{thm:guarantees} is in Appendix~\ref{app:guarantees}.

\begin{theorem}[Guarantees] \label{thm:guarantees}
    Given an input $\video$, a game model $\game(\network,\video,L^p,d)$, and an optimal strategy profile $\strategy = (\strategy_\one, \strategy_\two)$, the \emph{finite maximum safe radius} problem is to minimise the reward of initial state $s_0$ based on $\strategy$, i.e., 
    \begin{equation}
        \FMSR(\network, \flowset(\video), L^p, d, \magnitude) = \min \rewards(\strategy, s_0).
    \end{equation}
\end{theorem}

\section{Computation of the converging upper and lower bounds}

\correction{Since computing $\FMSR$ directly is difficult, we instead exploit algorithms to compute its upper and lower bounds. We emphasise that there is no approximation in the guarantees provided by the bounds, e.g., any adversarial perturbation of magnitude less than the lower bound will definitely not cause a misclassification. This includes all the possible inputs within the lower bound, not just a discrete subset.}

\subsection{Upper bound: gradient-based search}

We propose a gradient-based search algorithm to compute an upper bound of $\FMSR$. 
Here, we utilise the spatial features extracted from individual frames. 
\begin{definition}[Spatial Features] \label{dfn:spatial}
    Given a network $\network$, let $\network_\cnn$ denote the convolutional part, then $\network_\cnn: \video \mapsto \spatial \in \realNumber^{l \times m}$ maps from input $\video$ to its extracted \emph{spatial features} $\spatial$, which has consistent length $l$ of $\video$ and feature dimension $m$ of a frame. Then, we pass $\spatial$ into the recurrent part $\network_\rnn$ and obtain the classification, i.e., $\network_\rnn: \spatial \mapsto \network(\video, \class), \class \in \classes$.
\end{definition}
The objective is to manipulate optical flow as imperceptibly as possible while altering the final classification. 
We write the objective function as follows:
\begin{multline} \label{eqn:opObjective}
    \forall t \in [1, l-1], t \in \naturalNumber^+, \ \min \flow_t + \epsilon \cdot \gradient_{\flow_t} (\network, \video)
    \\
    \quad s.t. \quad
    \gradient_{\flow_t} (\network, \video) = 
    \dfrac{\partial \loss^{\network}_{\video}}{\partial \spatial} \odot \dfrac{\partial \spatial}{\partial \flow_t}
\end{multline}
where $\epsilon$ is a constant, and $\gradient_{\flow_t} (\network, \video)$ is the perturbation imposed on $\flow_t$. 
The key point is to minimise $\gradient_{\flow_t} (\network, \video)$ so that the perturbation is not noticeable while simultaneously changing $\network(\video)$. Here, we utilise the \emph{loss} of $\network$ on $\video$, denoted as $\loss^{\network}_{\video}$, to quantify the classification change. Intuitively, if $\loss^{\network}_{\video}$ increases, $\network(\video)$ is more likely to change. By utilising the concept of spatial features $\spatial$, we rewrite $\gradient_{\flow_t} (\network, \video)$ as $\partial \loss^{\network}_{\video} / \partial \spatial \odot \partial \spatial / \partial \flow_t$, where $\partial \loss^{\network}_{\video} / \partial \spatial$ denotes the gradient of the network's loss w.r.t. the spatial features, $\partial \spatial / \partial \flow_t$ denotes the gradient of the spatial features w.r.t. the optical flow, and $\odot$ denotes the Hadamard product.

We introduce the computation of the two parts below.
On one hand, $\partial \spatial / \partial \flow_t$ essentially exhibits the relation between spatial features and optical flow. Here we reuse input manipulation (Definition~\ref{dfn:manipulation}) to compute $\partial \spatial / \partial \flow_t$, though instead of manipulating the flow we impose perturbation directly on the frame. Intuitively, we manipulate the pixels of each frame to see how the subtle optical flow between the original and the manipulated frames will influence the spatial features. Each time we manipulate a single pixel of a frame, we get a new frame which is slightly different. If we perform $\manipulation_{\instruction,\magnitude}$ on pixel $\Frame[m,n]$, and denote the manipulated frame as $\Frame_{m,n}$, its spatial features as $\spatial_{m,n}$, the subtle optical flow between $\Frame_{m,n}$ and $\Frame$ as $\delta \flow_{m,n}$, then $\partial \spatial / \partial \flow_t$ can be computed as in Equation~(\ref{eqn:gradientSpatial}) below.
\begin{equation} \label{eqn:gradientSpatial}
    \dfrac{\partial \spatial}{\partial \flow_t} = 
    \begin{pmatrix}
        \dfrac{\distance{\spatial_{1,1}-\spatial}{p}}{\distance{\delta \flow_{1,1}}{p}} & \cdots & \dfrac{\distance{\spatial_{1,w}-\spatial}{p}}{\distance{\delta \flow_{1,w}}{p}} \\
        \vdots & \ddots & \vdots \\
        \dfrac{\distance{\spatial_{h,1}-\spatial}{p}}{\distance{\delta \flow_{h,1}}{p}} & \cdots & \dfrac{\distance{\spatial_{w,h}-\spatial}{p}}{\distance{\delta \flow_{w,h}}{p}}
    \end{pmatrix}
\end{equation}
On the other hand, $\partial \loss^{\network}_{\video} / \partial \spatial$ shows how the spatial features will influence the classification, which can be reflected by the loss of the network. After getting $\spatial$ from $\network_\cnn$, we can obtain $\loss^{\network}_{\video}$ from $\network_\rnn$. 
If we perform pixel manipulation $\manipulation_{\instruction,\magnitude} (\Frame[m,n])$ on frame $\Frame$, and obtain a new input, denoted as $\video_{\Frame[m,n]}$, then for this frame we have the gradient in Equation~(\ref{eqn:gradientLoss}) below.
\begin{equation} \label{eqn:gradientLoss}
    \dfrac{\partial \loss^{\network}_{\video}}{\partial \spatial} = 
    \begin{pmatrix}
        \dfrac{\loss^\network_{\video_{\Frame[1,1]}} - \loss^{\network}_{\video}}{\distance{\spatial_{1,1} - \spatial}{p}} & \cdots & \dfrac{\loss^\network_{\video_{\Frame[1,w]}} - \loss^{\network}_{\video}}{\distance{\spatial_{1,w} - \spatial}{p}} \\
        \vdots & \ddots & \vdots \\
        \dfrac{\loss^\network_{\video_{\Frame[h,1]}} - \loss^{\network}_{\video}}{\distance{\spatial_{h,1} - \spatial}{p}} & \cdots & \dfrac{\loss^\network_{\video_{\Frame[w,h]}} - \loss^{\network}_{\video}}{\distance{\spatial_{w,h} - \spatial}{p}}
    \end{pmatrix}
\end{equation}

\begin{remark}
    From Definition~\ref{dfn:spatial}, we know that the spatial features $\spatial = \network_\cnn(\video)$ only depend on each individual frame  $\Frame$ of $\video$ and do not capture the temporal information between frames.
    That is, when $\network_\cnn$ remains unchanged, $\spatial$ and $\Frame$ have a direct relation, which indicates that the gradient of the latter can reflect that of the former. Therefore, during implementation, instead of the distance between $\spatial_{m,n}$ and $\spatial$, we calculate that between $\Frame_{m,n}$ and $\Frame$, i.e., $\distance{\Frame_{m,n}-\Frame}{p}$.  
\end{remark}

\subsection{Lower bound: admissible A*}

\begin{algorithm}[t]
    \caption{Admissible A* for DNN Verification}
    \label{alg:admissibleA*}
    \small
    \SetAlgoLined
    \SetKwInOut{Input}{Input}
    \SetKwInOut{Output}{Output}
    \SetKwProg{Procedure}{procedure}{$:$}{}
    \SetKw{KwIn}{in}
    \SetInd{0.5em}{1.5em}
    \Input{\quad Game $\game(\network,\video,L^p,d)$, terminating condition $\terminate$}
    \Output{\quad Lower bound of $\FMSR$}
    \Procedure{\textsc{AdmissibleA*($\game(\network,\video,L^p,d),\terminate$)}}{
        $root \gets s_0$ \;
        \While{$(\neg \terminate)$}{
            $\flowset(root) \gets \playerone (root, \mathsf{Farneback})$ \;
            \For{$\flow_t$ \KwIn $\flowset(root)$}{
                $\flow_t[i] \gets \playertwo (\flow_t)$ \;
                $newnodes \gets \correction{\manipulation_{\atomic,\magnitude}(\flow_t)(i)}$ \;
                \For{$node$ \KwIn $newnodes$}{
                    $dist \gets \mathsf{DistanceEstimation}(node)$ \;}}
            $root \gets \mathsf{MaximumSafeRadius} (distances)$ \;}
        \Return $\distance{\flowset(root)-\flowset(s_0)}{p}$}
\end{algorithm}

We exploit admissible A* to compute a lower bound of $\playerone$'s reward, i.e., $\FMSR$. An A* algorithm gradually unfolds the game model into a tree, in the sense that it maintains a set of children nodes of the expanded partial tree, and computes an estimate for each node. The key point is that in each iteration it selects the node with the \emph{least} estimated value to expand. The estimation comprises two components: (1) the exact reward up to the current node, and (2) the estimated reward to reach the goal node. To guarantee the lower bound, we need to make sure that the estimated reward is minimal. For this part, we let the A* algorithm be \emph{admissible}, which means that, given a current node, it never overestimates the reward to the terminal goal state. 
For each state $s$ in the game model $\game$, we assign an estimated distance value
$\mathsf{Distance Estimation}(s) = \distance{\flowset(s)-\flowset(s_0)}{p} + heuristic(\flowset(s))$,
where $\distance{\flowset(s)-\flowset(s_0)}{p}$ is the distance from the original state $s_0$ to the current state $s$ based on the $L^p$ norm, and $heuristic(\flowset(s))$ is the admissible heuristic function that estimates the distance from the current state $s$ to the terminal state. Here, we use $\dist(L^p, \magnitude)$ in Equation~(\ref{eqn:bounds}). We present the admissible A* algorithm in Algorithm~\ref{alg:admissibleA*}.

\newcommand{\Mag}{\mathsf{magnitude}}
\newcommand{\Dir}{\mathsf{direction}}

\section{Experimental results}
\label{sec:Experiments}

This section presents the evaluation results of our framework to approximate the \emph{maximum safe radius w.r.t. optical flow} on a video dataset. We perform the experiments on a Linux server with NVIDIA GeForce GTX Titan Black GPUs, and the operating system is Ubuntu 14.04.3 LTS.
The results are obtained from a VGG16~\cite{simonyan2015very} + LSTM~\cite{hochreiter1997long} network on the  \emph{UCF101}~\cite{soomro2012ucf101} video dataset. Details about the dataset, the network structure, and training/testing parameters can be found in Appendix~\ref{app:network}.

\subsection{Adversarial examples via manipulating flows}

\begin{figure}[t]
    \centering
    \subfigure[$\mathsf{Soccer Juggling}$ at \SI{0}{\second} and \SI{1}{\second}.]{
    \includegraphics[width=0.32\linewidth]{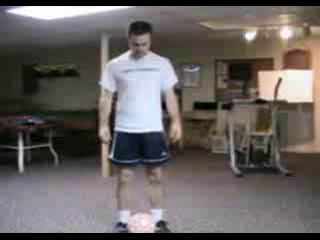}
    \includegraphics[width=0.32\linewidth]{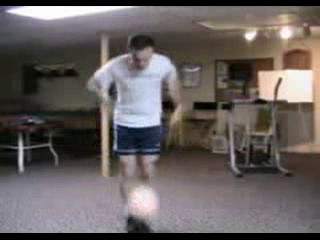}
    }
    \\[-1pt]
    \subfigure[Optical flow and its $\Mag$ (left) and $\Dir$ (right).]{    
    \includegraphics[width=0.32\linewidth]{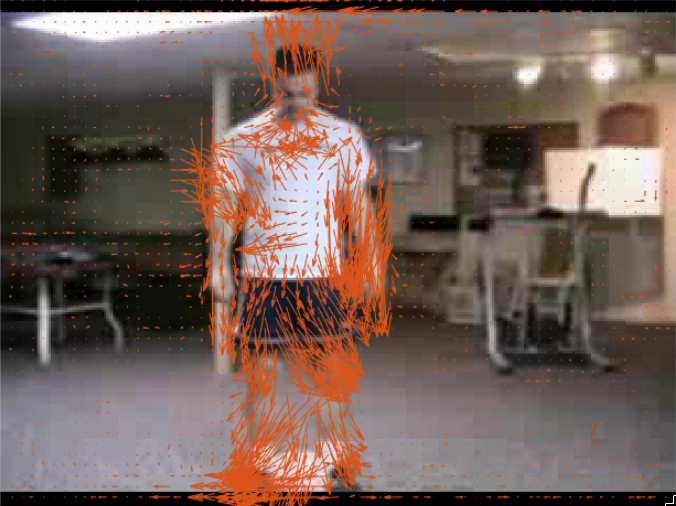}
    \includegraphics[width=0.32\linewidth]{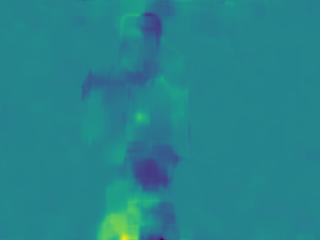}
    \includegraphics[width=0.32\linewidth]{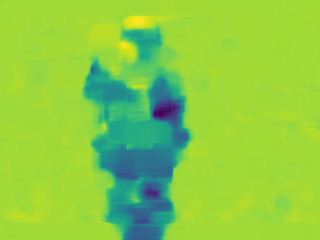}
    }
    \caption{Illustration of how optical flow is able to capture the dynamics of the moving objects. (a) Two sampled frames from $\mathsf{Soccer Juggling}$ with original size $320 \times 240 \times 3$. (b) The optical flow (red arrows) extracted between the frames, and its two characteristics: $\Mag$ and $\Dir$.}
    \label{fig:flow}
    \vspace{-1em}
\end{figure}
\begin{figure}[t]
    \centering
    \subfigure[Frame at \SI{1}{\second}.]{
    \includegraphics[width=0.23\linewidth]{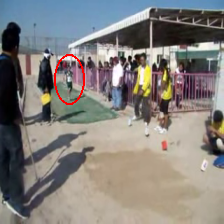}
    \label{}}
    \subfigure[$\Mag$: original, difference, and perturbed.]{
    \includegraphics[width=0.23\linewidth]{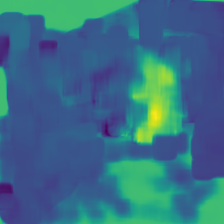}
    \includegraphics[width=0.23\linewidth]{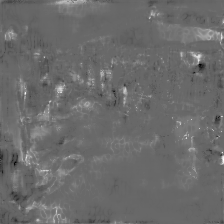}
    \includegraphics[width=0.23\linewidth]{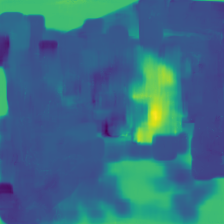}
    \label{}}
    \\[-5pt]
    \subfigure[Frame at \SI{2}{\second}.]{
    \includegraphics[width=0.23\linewidth]{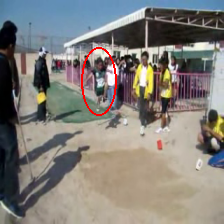}
    \label{}}
    \subfigure[$\Dir$: original, difference, and perturbed.]{
    \includegraphics[width=0.23\linewidth]{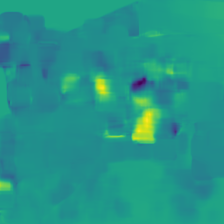}
    \includegraphics[width=0.23\linewidth]{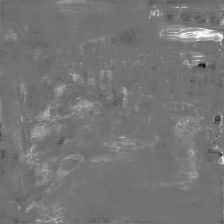}
    \includegraphics[width=0.23\linewidth]{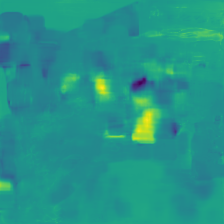}
    \label{}}
    \caption[Imperceptible perturbations on optical flow leading to misclassification.]{Imperceptible perturbations on optical flow, in terms of $\Mag$ and $\Dir$, leading to misclassification from $\mathsf{Long Jump}$ ($100.00\%$) to $\mathsf{Floor Gymnastics}$ ($86.10\%$). (a)(c) Sampled frames at \SI{1}{\second} and \SI{2}{\second} with size $224 \times 224 \times 3$. (b) Original and perturbed $\Mag$. (d) Original and perturbed $\Dir$.}
    \label{fig:adversary}
\end{figure}

We illustrate how \emph{optical flow} can capture the temporal dynamics of the moving objects in neighbouring frames. In this case, we exploit the Gunnar Farneb{\"a}ck algorithm~\cite{farneback2003two} as it computes the optical flow for all the pixels in a frame, i.e., \emph{dense} optical flow, instead of a sparse feature set. 
Figure~\ref{fig:flow} presents an optical flow generated from two adjacent frames of a video labelled as $\mathsf{Soccer Juggling}$: (a) shows two frames sampled at \SI{0}{\second} and \SI{1}{\second}; and (b) exhibits the characteristics of the flow: $\Mag$ and $\Dir$. 
We observe that, when the indoor background essentially remains unchanged, the motion of playing football is clearly captured by the flow.
More examples are included in Appendix~\ref{app:OpticalFlow}.

We now demonstrate how a very slight \emph{perturbation on the flow}, almost imperceptible to human eyes, can lead to a misclassification of the whole video. Figure~\ref{fig:adversary} exhibits that a video originally classified as $\mathsf{Long Jump}$ with confidence $100.00\%$ is manipulated into $\mathsf{Floor Gymnastics}$ with confidence $86.10\%$.
Two sampled frames at \SI{1}{\second} and \SI{2}{\second} are shown in the 1st column. If we compare the original optical flow of $\Mag$ and $\Dir$ (2nd column) generated from the frames with the perturbed ones (4th column), we can hardly notice the difference (3rd column). However, the classification of the video has changed.

\subsection{Converging upper and lower bounds}

\begin{figure}[t]
    \centering
    \includegraphics[width=0.24\linewidth]{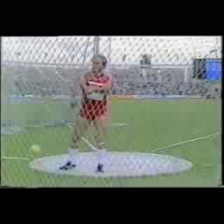}
    \includegraphics[width=0.24\linewidth]{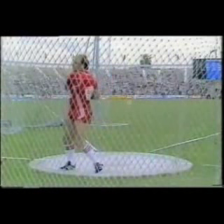}
    \includegraphics[width=0.24\linewidth]{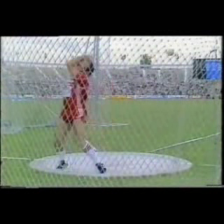}
    \includegraphics[width=0.24\linewidth]{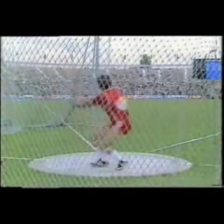}
    \\
    \includegraphics[width=0.24\linewidth]{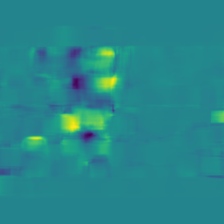}
    \includegraphics[width=0.24\linewidth]{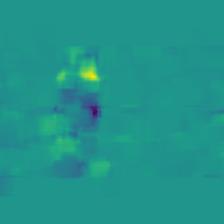}
    \includegraphics[width=0.24\linewidth]{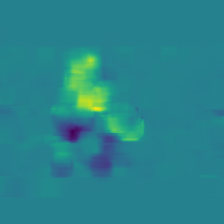}
    \includegraphics[width=0.24\linewidth]{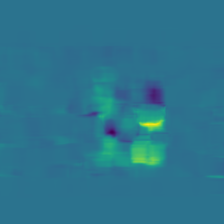}
    \\
    \includegraphics[width=0.24\linewidth]{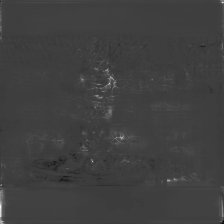}
    \includegraphics[width=0.24\linewidth]{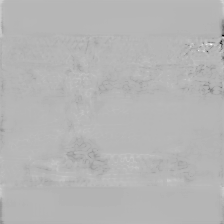}
    \includegraphics[width=0.24\linewidth]{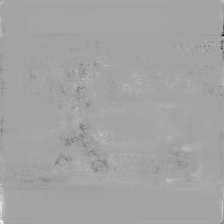}
    \includegraphics[width=0.24\linewidth]{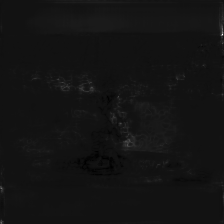}
    \\
    \includegraphics[width=0.24\linewidth]{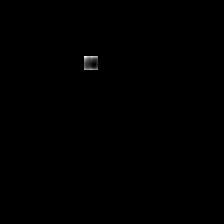}
    \includegraphics[width=0.24\linewidth]{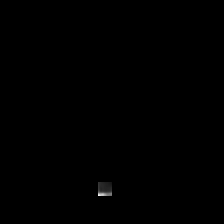}
    \includegraphics[width=0.24\linewidth]{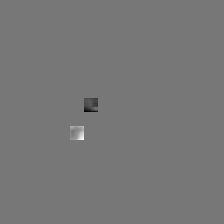}
    \includegraphics[width=0.24\linewidth]{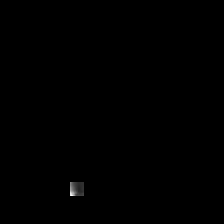}
    \caption{Examples of \emph{unsafe} and \emph{safe} perturbations on the optical flows of a $\mathsf{Hammer Throw}$ video. Top row: sampled frames from \SI{0}{\second} to \SI{3}{\second}. 2nd row: optical flows of the frames from \SI{0}{\second} to \SI{4}{\second}. 3rd row: \emph{unsafe} perturbations on the flows corresponding to the upper bound. Bottom row: \emph{safe} perturbations reflecting the lower bound.}
    \label{fig:perturbations_HammerThrow}
\end{figure}

\begin{figure}[t]
    \centering
    \includegraphics[width=1\linewidth]{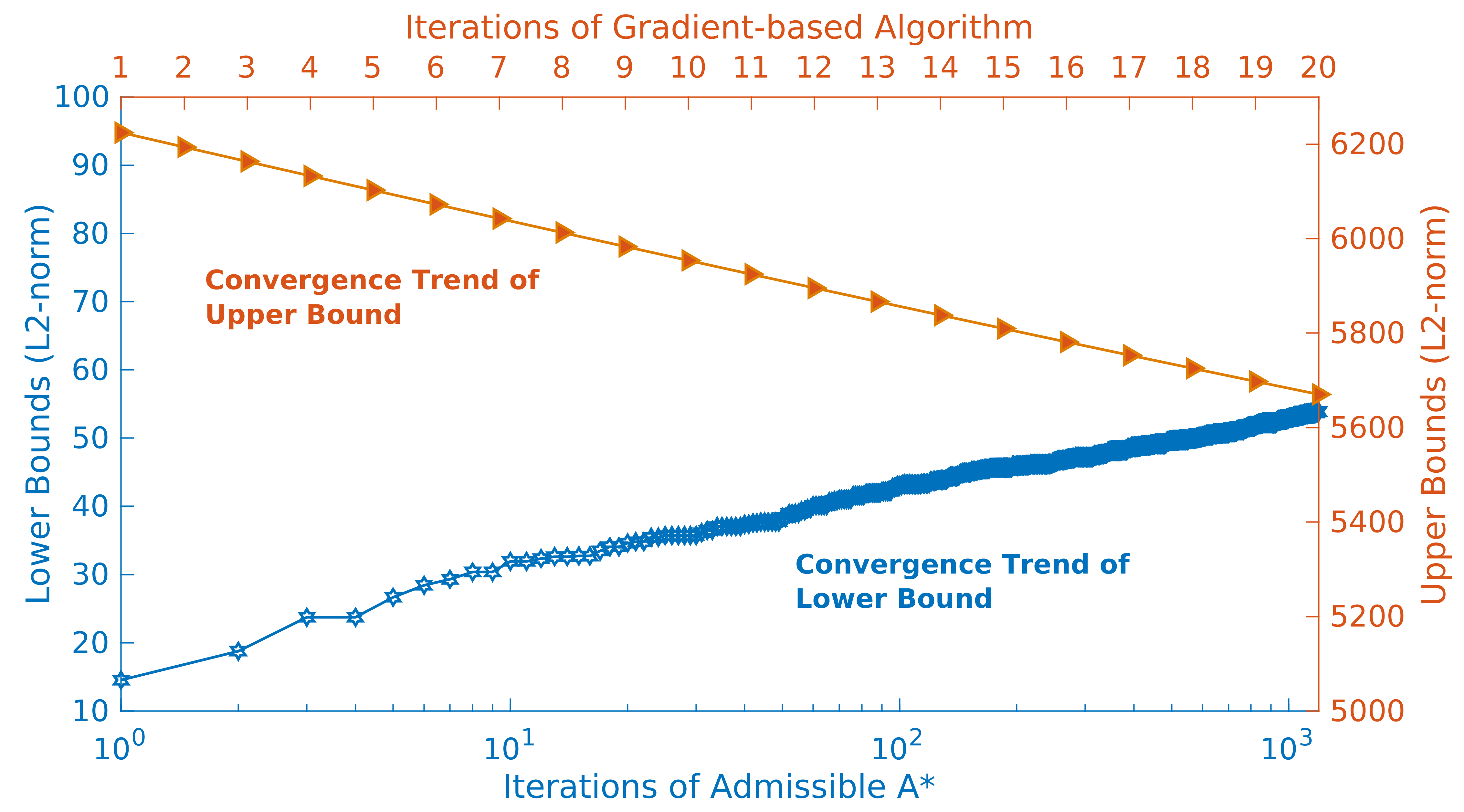}
    \caption{\emph{Converging bounds} of the maximum safe radius of the $\mathsf{Hammer Throw}$ video with respect to manipulations on extracted flows. 
    Decreasing \emph{upper} bound from the gradient-based algorithm shown in red, and increasing \emph{lower} bound from admissible A* in blue.
    \correction{Note that these two vertical axes have different scales.}}
    \label{fig:convergence_HammerThrow}
\end{figure}

We illustrate the convergence of the bound computation for the \emph{maximum safe radius} with respect to manipulations on the optical flows extracted from the consecutive frames of a video. Take a $\mathsf{Hammer Throw}$ video as an example. Figure~\ref{fig:perturbations_HammerThrow} exhibits four sampled frames (top row) and the optical flows extracted between them (2nd row). By utilising our framework, we present an approximation of $\MSR$ in Figure~\ref{fig:convergence_HammerThrow}, where the red line indicates the descending trend of the \emph{upper} bound, whereas the blue line denotes the ascending trend of the \emph{lower} bound. Intuitively, after $20$ iterations of the gradient-based algorithm, the upper bound, i.e., minimum distance to an adversarial example, is $5670.31$ based on the $L^2$ distance metric. \correction{That is, manipulations imposed on the flows exceeding this upper bound may be \emph{unsafe}}. Figure~\ref{fig:perturbations_HammerThrow} (3rd row) shows some of such unsafe perturbations on each optical flow, which result in the misclassification of the video into $\mathsf{Front Crawl}$ with confidence $99.86\%$. As for the lower bound, we observe that, after $1000$ iterations of the admissible A* algorithm, the lower bound reaches $52.95$. That is, manipulations within this $L^2$ norm ball are absolutely \emph{safe}. Some of such safe perturbations can be found in the bottom row of Figure~\ref{fig:perturbations_HammerThrow}.
Due to space limit, we include another example in Appendix~\ref{app:convergence}.

\subsection{Extension to naturally plausible distortions}

\begin{figure}[t]
    \centering
    \subfigure[Brightness increase and the corresponding unaffected optical flow.]{
        \centering
        \includegraphics[width=0.24\linewidth]{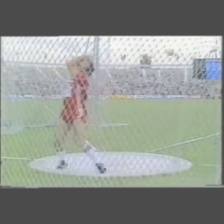}
        \includegraphics[width=0.24\linewidth]{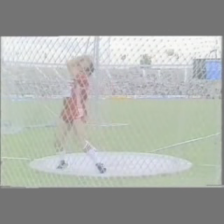}
        \includegraphics[width=0.24\linewidth]{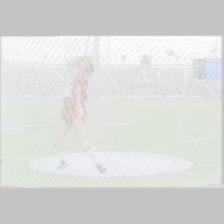}
        \includegraphics[width=0.24\linewidth]{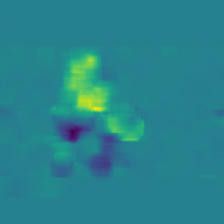}
        \label{fig:BrightnessFrames}}
    \\[-3pt]
    \subfigure[Lower bounds of the maximum safe radius.]{
        \centering
        \includegraphics[width=1\linewidth]{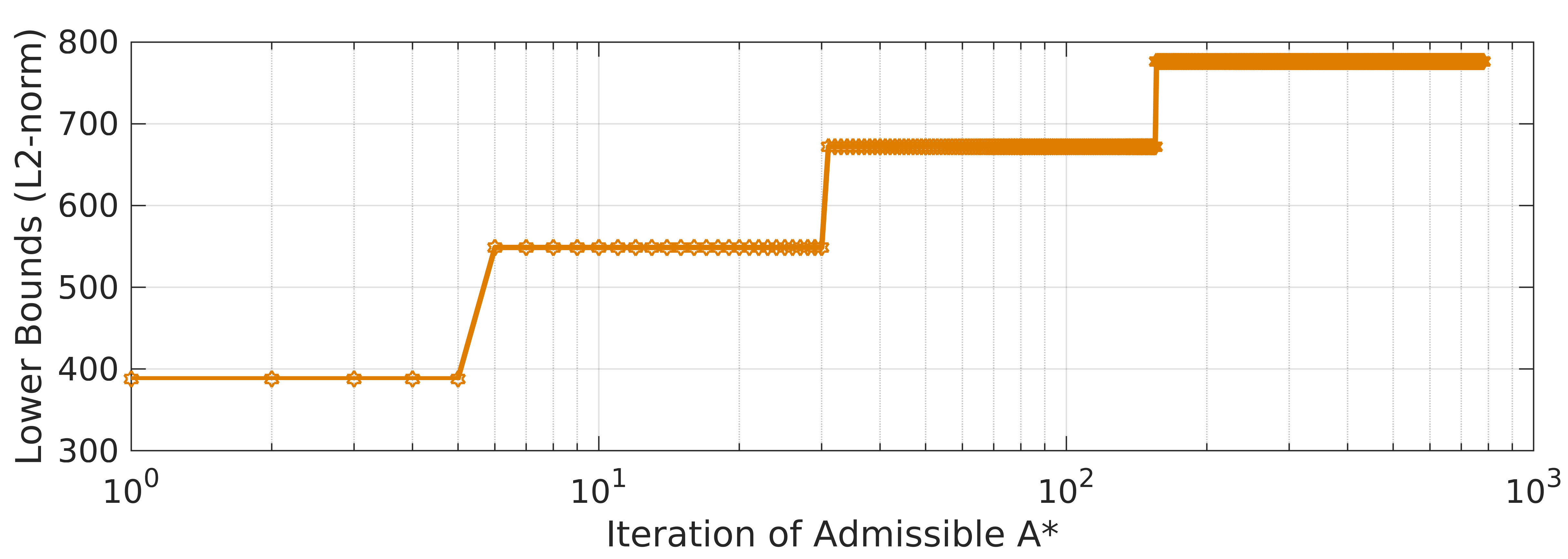}
        \label{fig:BrightnessLB}}
    \caption[Safe \emph{brightness} changes and the lower bounds.]{Safe \emph{brightness} changes to the $\mathsf{Hammer Throw}$ video and the corresponding lower bounds of the maximum safe radius. (a) The frame at \SI{2}{\second} of $\mathsf{Hammer Throw}$ with increasing brightness, and the optical flow extracted from the same frame taking into account the brightness change. (c) The ascending lower bounds of the maximum safe radius reflecting the brightness change.}
    \label{fig:BrightnessChange}
\end{figure}

\begin{figure}[t]
    \centering
        \includegraphics[width=0.24\linewidth]{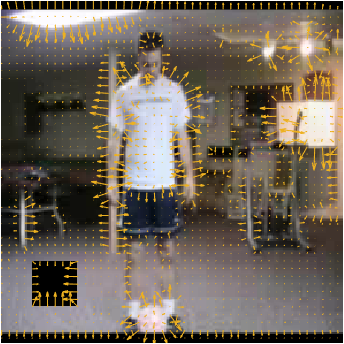}
        \includegraphics[width=0.24\linewidth]{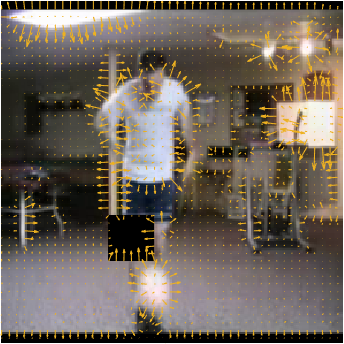}
        \includegraphics[width=0.24\linewidth]{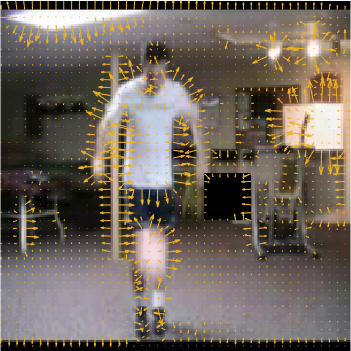}
        \includegraphics[width=0.24\linewidth]{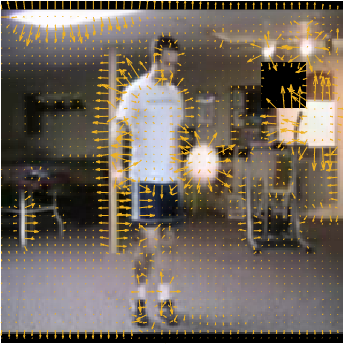}
        \\
        \includegraphics[width=0.24\linewidth]{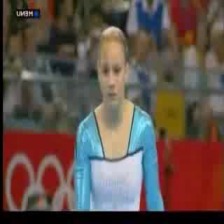}
        \includegraphics[width=0.24\linewidth]{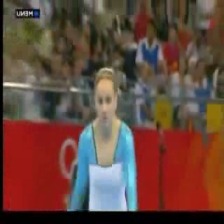}
        \includegraphics[width=0.24\linewidth]{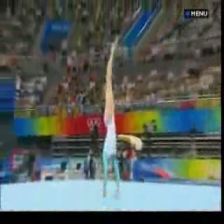}
        \includegraphics[width=0.24\linewidth]{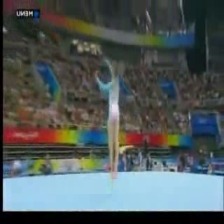}
        \\
        \includegraphics[width=0.24\linewidth]{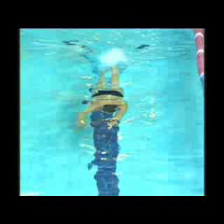}
        \includegraphics[width=0.24\linewidth]{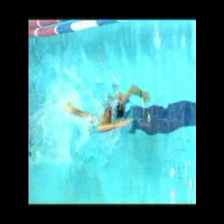}
        \includegraphics[width=0.24\linewidth]{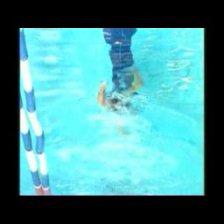}
        \includegraphics[width=0.24\linewidth]{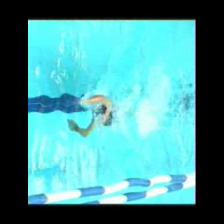}
    \caption[Possible extension to more \emph{naturally plausible distortions}.]{Some possible extensions of the adversarial perturbations to more \emph{naturally plausible distortions}. Top: ``camera occlusion'' to the $\mathsf{Soccer Juggling}$ video with the \emph{Horn-Schunck} optical flow method. Middle: ``horizontal flip'' to the $\mathsf{Floor Gymnastics}$ video. Bottom: ``angular rotation'' to the $\mathsf{Front Crawl}$ video.}
    \label{fig:NaturalDistortions}
\end{figure}

\correction{Our framework can be extended to practical applications where distortions are more \emph{natural and physically plausible} to the modality of the data itself, because all the perturbations preserving the semantic content of a video are essentially compositions of various atomic manipulations, and thus can be easily incorporated.}

Take the ``\emph{brightness change}'' perturbation as an example. As illustrated in Figure~\ref{fig:BrightnessChange}, we increase the brightness of the $\mathsf{Hammer Throw}$ video on the frame level. That is, each pixel in the same frame is simultaneously brightened by the atomic manipulation $\magnitude$, thus resulting in the overall distance to the original video increasing by $\dist'(L^p, \magnitude \cdot w \cdot h)$, where $w$ denotes the width of the frame and $h$ height. The corresponding lower bounds of $\MSR$ are computed in Figure~\ref{fig:BrightnessLB}. Intuitively, this means that any degree of brightness alteration is definitely \emph{safe} as long as the distance to the original video is less than the computed lower bound. For instance, after $10$ iterations, the lower bound is \num{548.68} based on the $L^2$ norm, then any frame-level brightness increase less than \num{548.68} in the Euclidean distance will not change the classification of this video.
One interesting phenomenon observed is that, as exhibited in Figure~\ref{fig:BrightnessFrames}, when the brightness of a frame increases, the extracted optical flow on the same frame is \emph{not} significantly affected, due to the fact that the motion is relatively unchanged. In other words, optical flow can naturally discard some perturbations that do not alter the underlying temporal dynamics.

Apart from the ``brightness change'', we include some other possible natural distortions of the adversarial perturbations in Figure~\ref{fig:NaturalDistortions}.
We observe that the ``camera occlusion'' here is very similar to the safe perturbations of the $\mathsf{Hammer Throw}$ video in Figure~\ref{fig:perturbations_HammerThrow} (bottom row), and thus can be handled using similar methods. 
The ``horizontal flip'' and the ``angular rotation'' involve manipulations that form a group, and to deal with those our approach would need to be extended, for example by incorporating network invariances. 
Finally, regarding these various adversarial perturbations, we remark that whether the perturbations are visible for a human largely depends on the \emph{manipulation} type -- with the same distance to the original input, manipulations such as these physically plausible distortions are certainly more visible than unsafe perturbations produced by the gradient-based search algorithm and the safe perturbations created from the admissible A* algorithm.

\subsection{Efficiency and scalability}

\begin{figure}[t]
    \centering
    \includegraphics[width=1\linewidth]{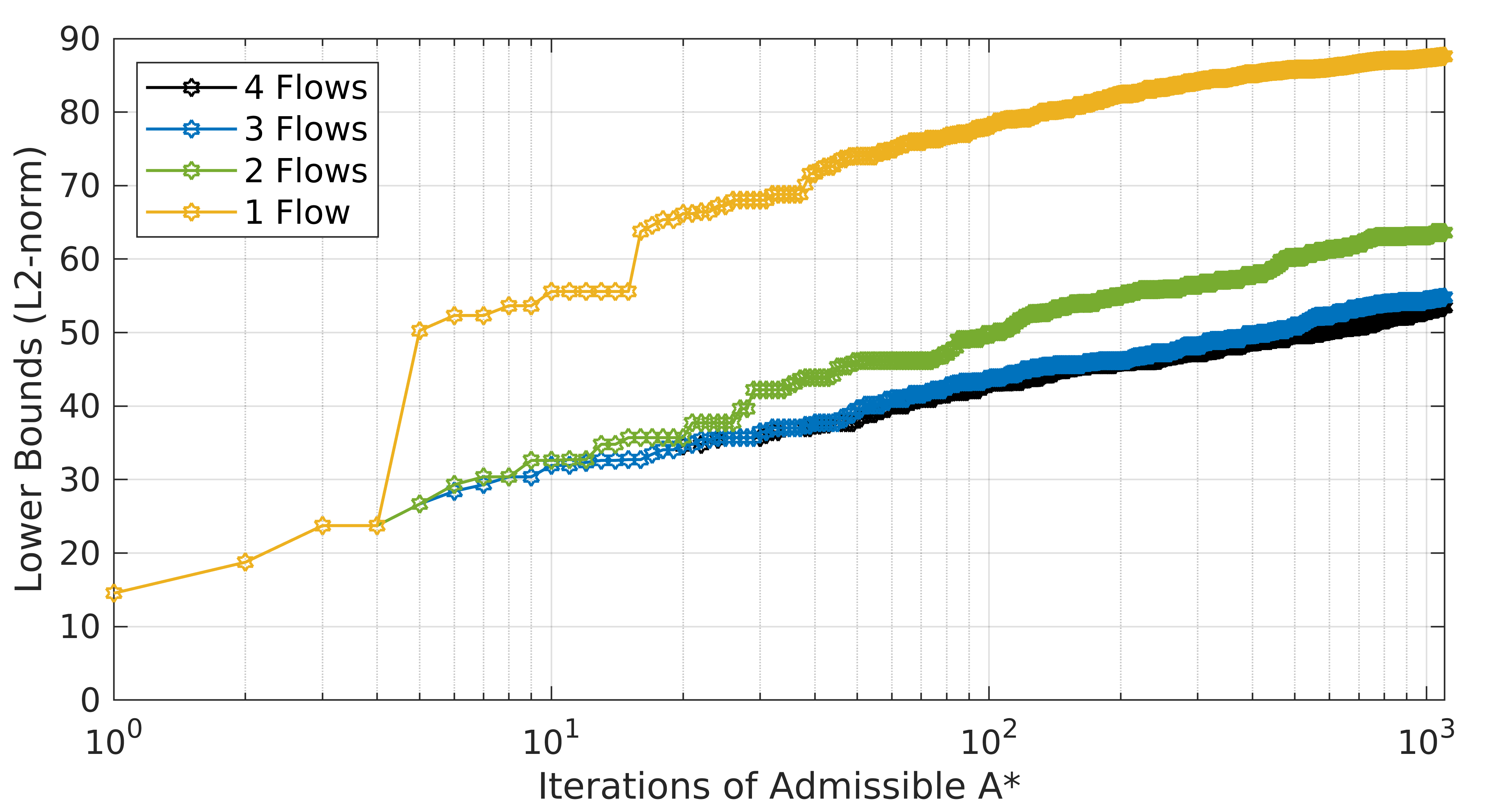}
    \caption{Different lower bounds of the maximum safe radius of a $\mathsf{Hammer Throw}$ video with varying number of manipulated flows.}
    \label{fig:scalability}
\end{figure}

As for the \emph{computation time}, the upper bound requires the gradient of optical flow with respect to the frame, and because we extract dense optical flow, the algorithm needs to traverse each pixel of a frame to impose atomic manipulations; thus it takes around $30$ minutes to retrieve the gradient of each frame. Once the gradient of the whole video is obtained, and the framework enters into the cooperative game, i.e., the expansion of the tree, each iteration takes minutes. Meanwhile, for the lower bound, the admissible A* algorithm expands the game tree in each iteration which takes minutes, and updates the lower bound wherever applicable. Note that initially the lower bound may be updated in each iteration, but when the size of the game tree increases, it can take hours to update.
Moreover, we analyse the \emph{scalability} of our framework via an example of a $\mathsf{Hammer Throw}$ video in Figure~\ref{fig:scalability}, which shows the lower bounds obtained with respect to varying dimensions of the manipulated flows. 
We observe that, within the same number of iterations, manipulating fewer flows leads to faster convergence, especially when the flows contain more spatial and temporal features.

\section{Conclusion}

In this work, we study the \emph{maximum safe radius} problem for neural networks, including CNNs and RNNs, with respect to the optical flow extracted from sequential videos. 
\correction{Possible future work includes extending our framework to provide robustness guarantees for application domains where time-series input data and recurrent units are also involved, for example natural language processing.}

\noindent {\bf Acknowledgements:} 
This project has received funding from the European Research
20 Council (ERC) under the European Union’s Horizon 2020 research and innovation programme (grant agreement No. 834115) and the EPSRC Programme Grant on Mobile Autonomy (EP/M019918/1).

\newpage
{
\small
\bibliographystyle{ieee_fullname}
\bibliography{main}
}

\newpage
\appendix

\newcommand{\confidence}{\mathsf{Mar}}

\section{Appendix}
\label{sec:Appendix}

\subsection{Proof of the error bounds in Theorem~\ref{thm:bounds}}
\label{app:error}

In this section, we provide detailed proof for the error bounds in Theorem~\ref{thm:bounds}, in particular, regarding the value of $\dist'(L^p, \magnitude)$ in Equation~(\ref{eqn:Lipschitz}).

\begin{proof}
    We first define the concept of the \emph{minimum confidence margin}.
    \begin{definition}[Minimum Confidence Margin]
        Given a network $\network$, an input $\video$, and a class $\class$, we define the \emph{minimum confidence margin} as
        \begin{equation}
            \confidence(\video, \class) = \min_{\class' \in \classes, \class' \neq \class} 
            \{ \network(\video, \class) - \network(\video, \class') \}.
        \end{equation}
    \end{definition}
    Intuitively, the minimum confidence margin is the discrepancy between the maximum confidence of $\video$ being classified as $\class$ and the next largest confidence of $\video$ being classified as $\class'$.
    Then, for any input $\video'$ whose optical flow \correction{sequence} is in the subspace of a grid point $\gridPoint$, and the input $\video$ corresponding to this optical flow \correction{sequence} $\gridPoint$, we have
        \begin{equation}
        \begin{split}
          & \confidence(\video,\network(\video)) - \confidence(\video',\network(\video)) \\
        = & \min_{\class \in \classes, \class \neq \network(\video)} \{\network(\video,\network(\video)) - \network(\video, \class)\} \\
        & \quad \quad \quad \quad \quad \quad
        - \min_{\class \in \classes, \class \neq \network(\video)} \{\network(\video',\network(\video)) - \network(\video',\class)\} \\
        \leq & \max_{\class \in \classes, \class \neq \network(\video)} \{\network(\video,\network(\video)) -  \network(\video,\class) \\
        & \quad \quad \quad \quad \quad \quad \quad \quad \quad \quad \quad
        - \network(\video',\network(\video)) + \network(\video',\class)\} \\
        \leq & \max_{\class \in \classes, \class \neq \network(\video)} \{\abs{\network(\video, \network(\video)) - \network(\video',\network(\video))} \\
        & \quad \quad \quad \quad \quad \quad \quad \quad \quad \quad \quad \quad \ \
        + \abs{\network(\video',\class) - \network(\video,\class)} \} \\
        \leq & \max_{\class \in \classes, \class \neq \network(\video)} \Lipschitz_{\network(\video)} \cdot \distance{\video-\video'}{p} + \Lipschitz_{\class} \cdot \distance{\video-\video'}{p} \\
        \leq & \max_{\class \in \classes, \class \neq \network(\video)} (\Lipschitz_{\network(\video)} + \Lipschitz_{\class}) \cdot \distance{\video-\video'}{p} \\
        \leq & \max_{\class \in \classes, \class \neq \network(\video)} (\Lipschitz_{\network(\video)}+\Lipschitz_{\class}) \cdot \dist'(L^p,\tau)
        \end{split}
    \end{equation}
    Now, since the optical flow \correction{sequence} of $\video'$ is in the subspace of $\gridPoint$, we need to ensure that no class change occurs between $\video$ and $\video'$. That is, $\confidence(\video',\network(\video)) \geq 0$, which means $\confidence(\video,\network(\video)) - \confidence(\video',\network(\video)) \leq \confidence(\video,\network(\video))$. Therefore, we have
    \begin{equation}
        \max_{\class \in \classes, \class \neq \network(\video)} (\Lipschitz_{\network(\video)}+\Lipschitz_{\class}) \cdot \dist'(L^p,\tau)
        \leq \confidence(\video,\network(\video)).
    \end{equation}
    And as $\gridPoint$ is a grid point, the minimum confidence margin for its corresponding input $\video$ can be computed. Finally, we replace $\confidence(\video,\network(\video))$ with its definition, then we have
    \begin{equation}
        \dist'(L^p, \magnitude) \leq 
        \dfrac{\displaystyle \min_{\class \in \classes, \class \neq \network(\video)}
        \{ \network(\video, \network(\video)) - \network(\video, \class) \} }
        {\displaystyle \max_{\class \in \classes, \class \neq \network(\video)} 
        (\Lipschitz_{\network(\video)} + \Lipschitz_\class) }.
    \end{equation}
\end{proof}

\subsection{Proof of the guarantees in Theorem~\ref{thm:guarantees}}
\label{app:guarantees}

In this section, we provide a detailed proof for the robustness guarantees in Theorem~\ref{thm:guarantees}.

\begin{proof}
    On one hand, we show that $\distance{\flowset(\video')-\flowset(\video)}{p} \geq \rewards(\strategy, s_0)$ for any optical flow \correction{sequence} $\flowset(\video')$ as a $\magnitude$-grid point, such that $\flowset(\video') \in \ball (\flowset(\video), L^p, d)$ and its corresponding input is an adversarial example. Intuitively, it means that $\playerone$'s reward from the game $\game$ in the initial state $s_0$ is no greater than the $L^p$ distance to any $\magnitude$-grid manipulated optical flow \correction{sequence}. That is, the reward value $\rewards(\strategy, s_0)$, once computed, is a lower bound of the optimisation problem $\FMSR (\network, \flowset(\video), L^p, d, \magnitude)$. Note that the reward value can be obtained as every $\magnitude$-grid point can be reached by some game play, i.e., a sequence of atomic manipulations.
    
    On the other hand, from the termination condition $\terminate(\Path)$ of the game, we observe that, for some $\flowset(\video')$, if $\rewards(\strategy, s_0) \leq \distance{\flowset(\video')-\flowset(\video)}{p}$ holds, then there must exist some other $\flowset(\video'')$ such that $\rewards(\strategy, s_0) = \distance{\flowset(\video''-\flowset(\video))}{p}$. Therefore, we have that $\rewards(\strategy, s_0)$ is the minimum value of $\distance{\flowset(\video''-\flowset(\video))}{p}$ among all the $\magnitude$-grid points $\flowset(\video')$ such that $\flowset(\video') \in \ball (\flowset(\video), L^p, d)$ and their corresponding inputs are adversarial examples.
    
    Finally, we observe that the minimum value of $\distance{\flowset(\video')-\flowset(\video)}{p}$ is equivalent to the optical flow value required by Equation~(\ref{eqn:FMSR}).
\end{proof}

\subsection{Details of the video dataset and the network}
\label{app:network}

As a popular benchmark for human action recognition in videos, \emph{UCF101}~\cite{soomro2012ucf101} consists of 101 annotated action classes, e.g., $\mathsf{Juggling Balls}$ (human-object interaction), $\mathsf{Handstand Pushups}$ (body-motion only), $\mathsf{Hair Cut}$ (human-human interaction), $\mathsf{Playing Piano}$ (playing musical instruments), and $\mathsf{Floor Gymnastics}$ (sports).
It labels \num{13320} video clips of $27$ hours in total, and each frame has dimension $320 \times 240 \times 3$.

In the experiments, we exploit a VGG16 + LSTM architecture, in the sense of utilising the \emph{VGG16} network to extract the spatial features from the UCF101 video dataset and then passing these features to a separate RNN unit \emph{LSTM}. 
For each video, we sample a frame every \SI{1000}{\ms} and stitch them together into a sequence of frames. Specifically, we run every frame from every video through VGG16 with input size $224 \times 224 \times 3$, excluding the top classification part of the network, i.e., saving the output from the final Max-Pooling layer. Hence, for each video, we retrieve a sequence of extracted spatial features. Subsequently, we pass the features into a single LSTM layer, followed by a Dense layer with some Dropout in between. Eventually, after the final Dense layer with activation function Softmax, we get the classification outcome.

We use the $\mathsf{categorical \ cross \text{-} entropy}$ loss function and the $\mathsf{accuracy}$ metrics for both the VGG16 and LSTM models. Whilst the former has a $\mathsf{SGD}$ optimiser and directly exploits the $\mathsf{imagenet}$ weights, we train the latter through a $\mathsf{rmsprop}$ optimiser and get $99.15\%$ training accuracy as well as $99.72\%$ testing accuracy. Specifically, when the \emph{loss} difference cannot reflect the subtle perturbation on optical flow during the computation of upper bounds, we use the discrepancy of $\mathsf{logit}$ values instead.

\subsection{More examples of the optical flows extracted from different videos}
\label{app:OpticalFlow}

\begin{figure}[t]
    \centering
    \includegraphics[width=0.24\linewidth]{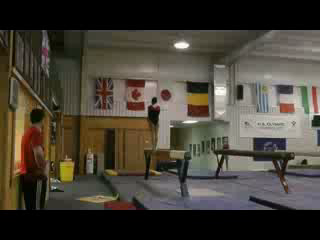}
    \includegraphics[width=0.24\linewidth]{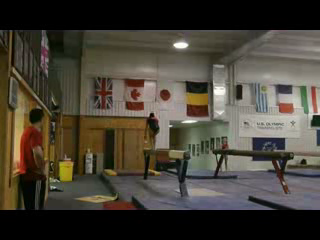}
    \includegraphics[width=0.24\linewidth]{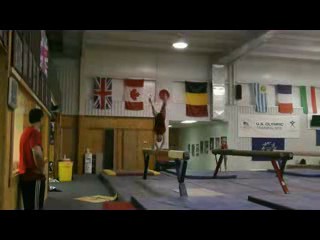}
    \includegraphics[width=0.24\linewidth]{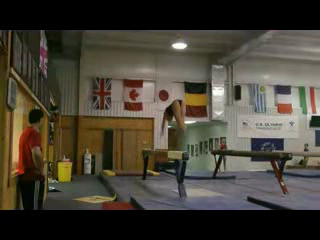}
    \\
    \includegraphics[width=0.24\linewidth]{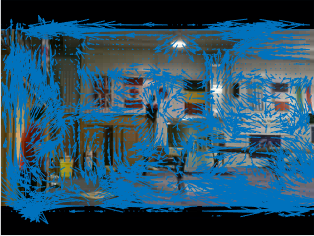}
    \includegraphics[width=0.24\linewidth]{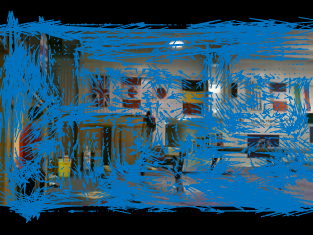}
    \includegraphics[width=0.24\linewidth]{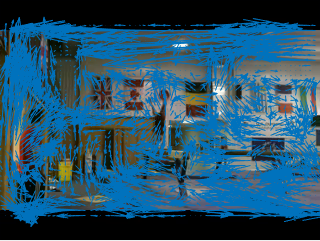}
    \includegraphics[width=0.24\linewidth]{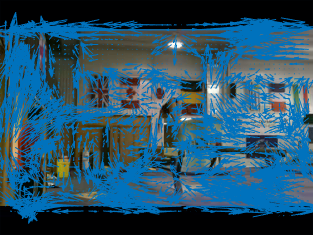}
    \\
    \includegraphics[width=0.24\linewidth]{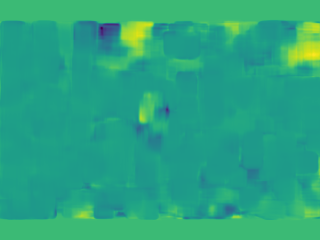}
    \includegraphics[width=0.24\linewidth]{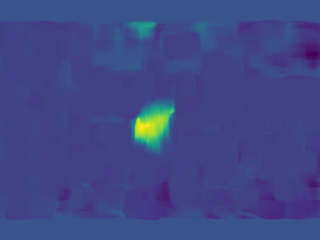}
    \includegraphics[width=0.24\linewidth]{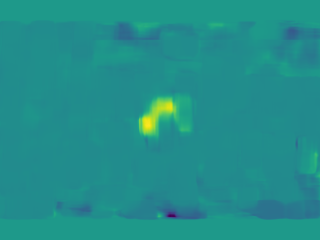}
    \includegraphics[width=0.24\linewidth]{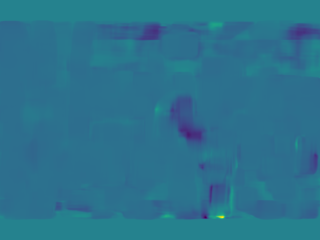}
    \\
    \includegraphics[width=0.24\linewidth]{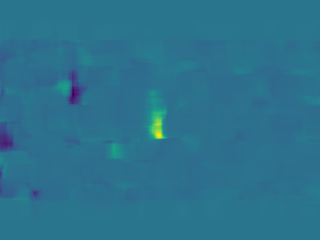}
    \includegraphics[width=0.24\linewidth]{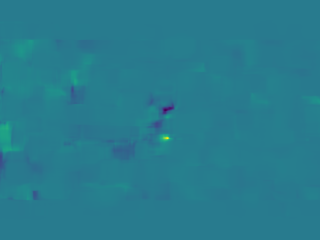}
    \includegraphics[width=0.24\linewidth]{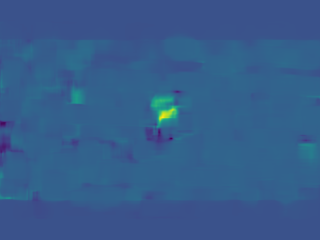}
    \includegraphics[width=0.24\linewidth]{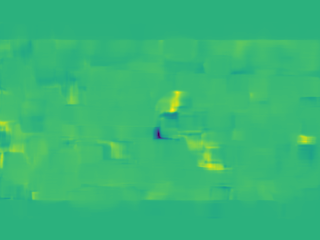}
    \caption{Examples of the optical flows extracted from a $\mathsf{Balance Beam}$ video. Top row: four sampled frames from \SI{0}{\second} to \SI{3}{\second} with original size $320 \times 240 \times 3$. 2nd row: the optical flows (blue arrows) extracted between the frames. 3rd row: one of optical flow's characteristics: $\Mag$. Bottom row: the other optical flow characteristics: $\Dir$.}
    \label{fig:OpticalFlow_BalanceBeam}
\end{figure}

\begin{figure}[t]
    \centering
    \includegraphics[width=0.24\linewidth]{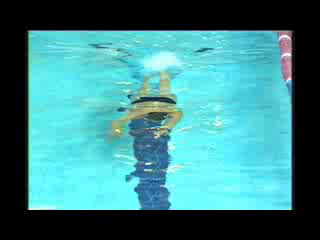}
    \includegraphics[width=0.24\linewidth]{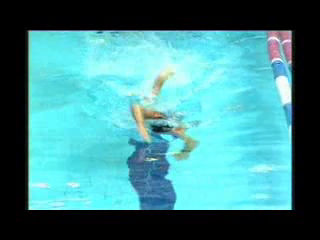}
    \includegraphics[width=0.24\linewidth]{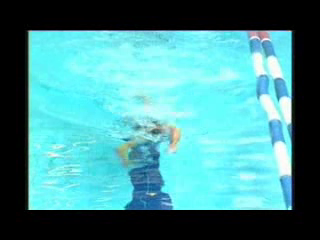}
    \includegraphics[width=0.24\linewidth]{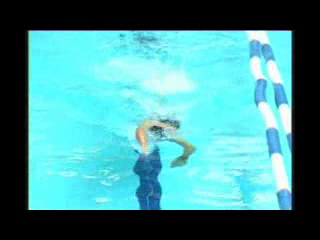}
    \\
    \includegraphics[width=0.24\linewidth]{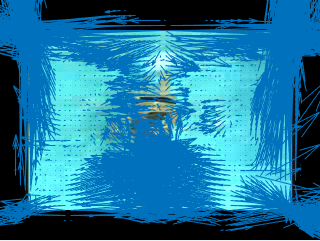}
    \includegraphics[width=0.24\linewidth]{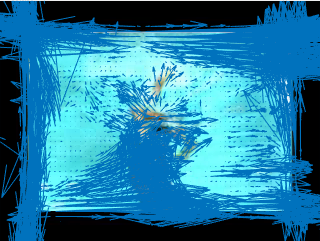}
    \includegraphics[width=0.24\linewidth]{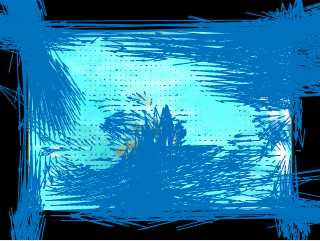}
    \includegraphics[width=0.24\linewidth]{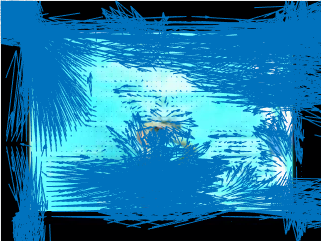}
    \\
    \includegraphics[width=0.24\linewidth]{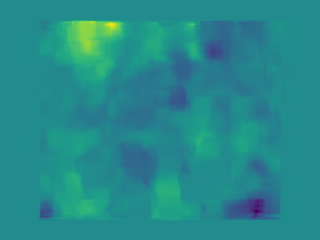}
    \includegraphics[width=0.24\linewidth]{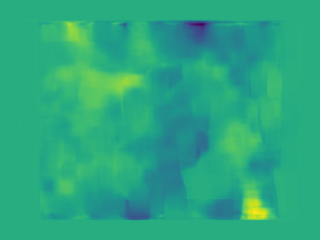}
    \includegraphics[width=0.24\linewidth]{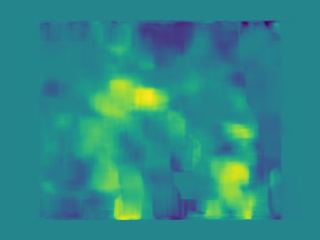}
    \includegraphics[width=0.24\linewidth]{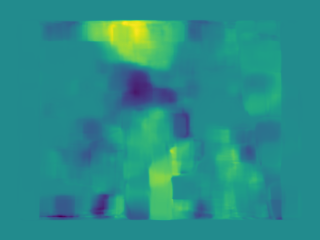}
    \\
    \includegraphics[width=0.24\linewidth]{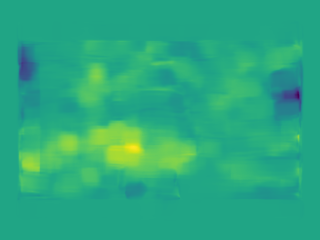}
    \includegraphics[width=0.24\linewidth]{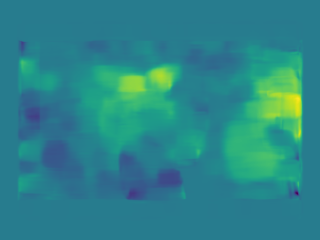}
    \includegraphics[width=0.24\linewidth]{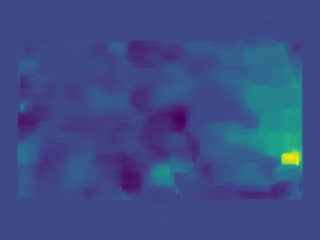}
    \includegraphics[width=0.24\linewidth]{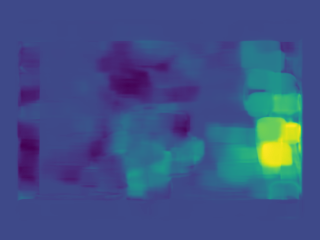}
    \caption{Examples of the optical flows extracted from a $\mathsf{Front Crawl}$ video. Top row: four sampled frames from \SI{0}{\second} to \SI{3}{\second} with original size $320 \times 240 \times 3$. 2nd row: the optical flows (blue arrows) extracted between the frames. 3rd row: one of optical flow's characteristics: $\Mag$. Bottom row: the other optical flow characteristics: $\Dir$.}
    \label{fig:OpticalFlow_FrontCrawl}
\end{figure}

Apart from Figure~\ref{fig:flow} in Section~\ref{sec:Experiments}, here we include more examples of the optical flows extracted from another two videos with classifications $\mathsf{Balance Beam}$ (Figure~\ref{fig:OpticalFlow_BalanceBeam}) and $\mathsf{Front Crawl}$ (Figure~\ref{fig:OpticalFlow_FrontCrawl}).


\subsection{Another example of the converging upper and lower bounds}
\label{app:convergence}

\begin{figure}[t]
    \centering
    \includegraphics[width=0.19\linewidth]{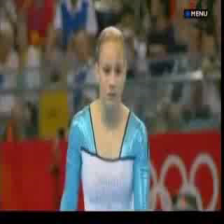}
    \includegraphics[width=0.19\linewidth]{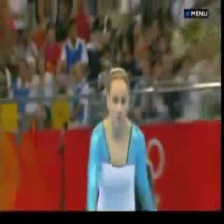}
    \includegraphics[width=0.19\linewidth]{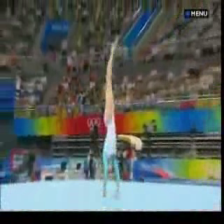}
    \includegraphics[width=0.19\linewidth]{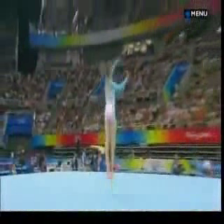}
    \includegraphics[width=0.19\linewidth]{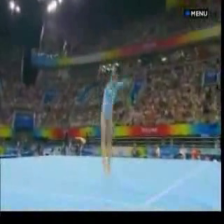}
    \\
    \includegraphics[width=0.19\linewidth]{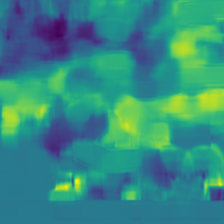}
    \includegraphics[width=0.19\linewidth]{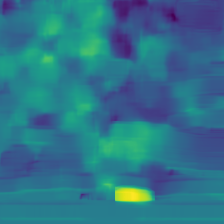}
    \includegraphics[width=0.19\linewidth]{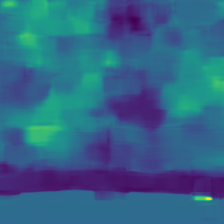}
    \includegraphics[width=0.19\linewidth]{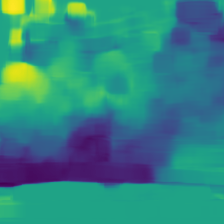}
    \includegraphics[width=0.19\linewidth]{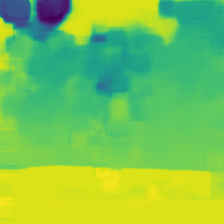}
    \\
    \includegraphics[width=0.19\linewidth]{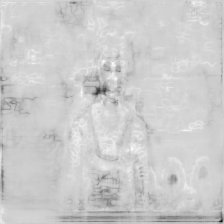}
    \includegraphics[width=0.19\linewidth]{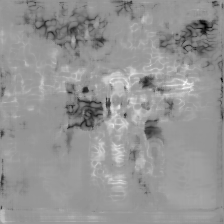}
    \includegraphics[width=0.19\linewidth]{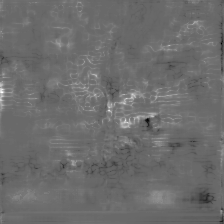}
    \includegraphics[width=0.19\linewidth]{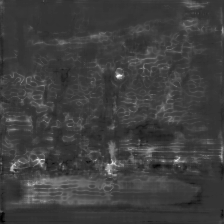}
    \includegraphics[width=0.19\linewidth]{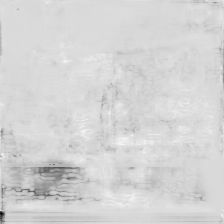}
    \\
    \includegraphics[width=0.19\linewidth]{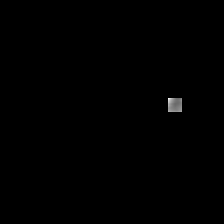}
    \includegraphics[width=0.19\linewidth]{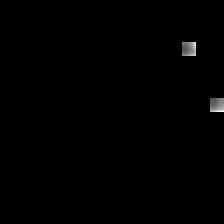}
    \includegraphics[width=0.19\linewidth]{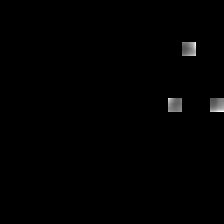}
    \includegraphics[width=0.19\linewidth]{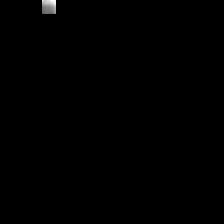}
    \includegraphics[width=0.19\linewidth]{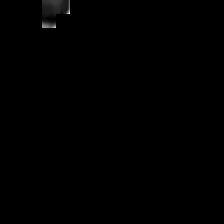}
    \caption{Examples of \emph{unsafe} and \emph{safe} perturbations on the optical flows of a $\mathsf{Floor Gymnastics}$ video. Top row: five sampled frames from \SI{0}{\second} to \SI{4}{\second}. 2nd row: optical flows of the frames from \SI{0}{\second} to \SI{5}{\second}. 3rd row: \emph{unsafe} perturbations on the flows corresponding to the upper bound. Bottom row: \emph{safe} perturbations on the flows corresponding to the lower bound.}
    \label{fig:perturbations_FloorGymnastics}
\end{figure}

\begin{figure}[t]
    \centering
    \includegraphics[width=1\linewidth]{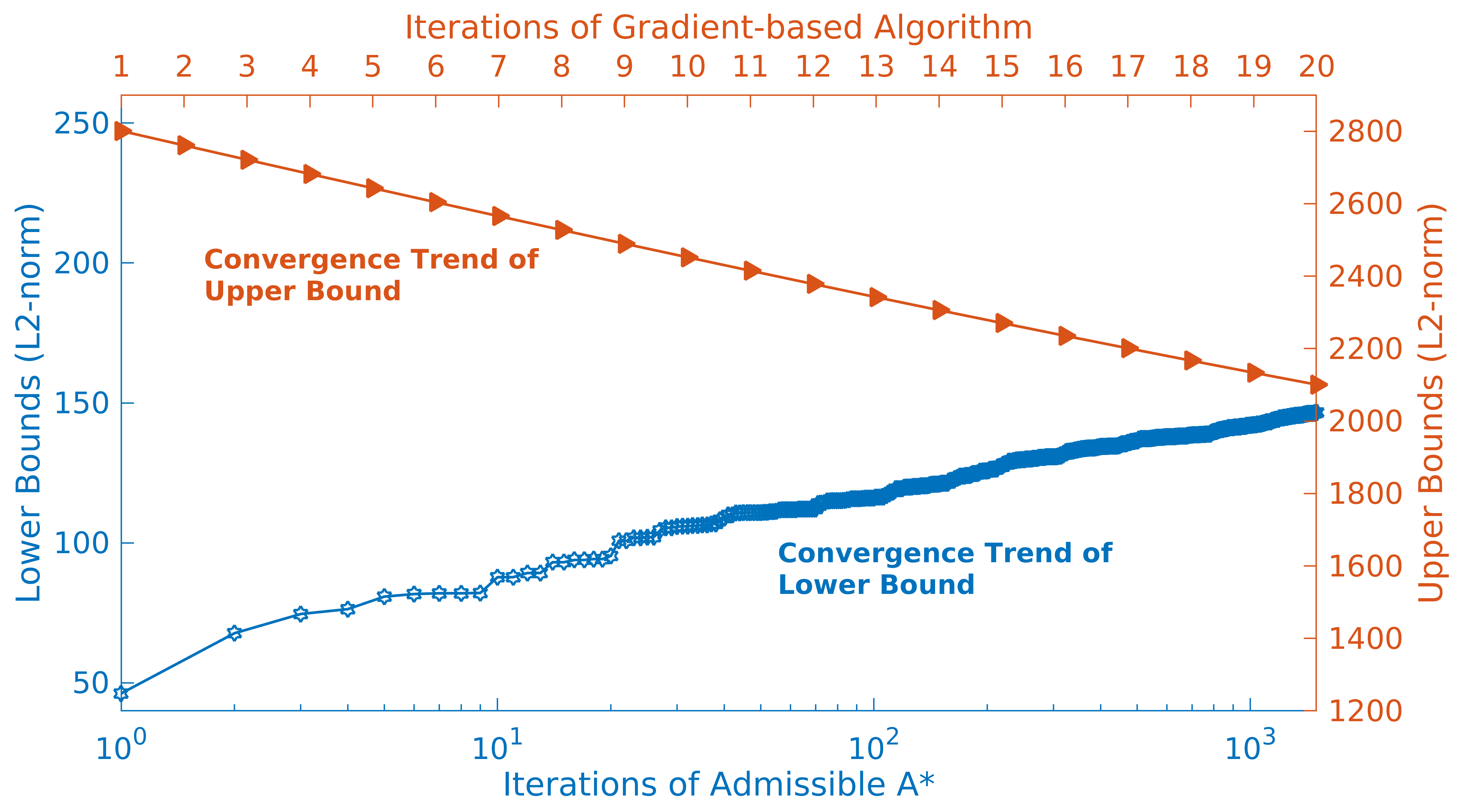}
    \caption{\emph{Converging bounds} of the maximum safe radius of the $\mathsf{Floor Gymnastics}$ video with respect to manipulations on extracted optical flows. The red line denotes the decreasing \emph{upper} bound from the gradient-based algorithm, and the blue line denotes the increasing \emph{lower} bound from admissible A*.
    }
    \label{fig:convergence_FloorGymnastics}
\end{figure}

Apart from the $\mathsf{Hammer Throw}$ example (Figures~\ref{fig:perturbations_HammerThrow} and \ref{fig:convergence_HammerThrow}, Section~\ref{sec:Experiments}), we include another example to illustrate the convergence of the upper and lower bounds. Similarly, Figure~\ref{fig:perturbations_FloorGymnastics} exhibits five sampled frames (top row) from a $\mathsf{Floor Gymnastics}$ video and the optical flows extracted between them (2nd row). The descending upper bounds (red) and the ascending lower bounds (blue) to approximate the value of $\MSR$ are presented in Figure~\ref{fig:convergence_FloorGymnastics}. Intuitively, after $20$ iterations of the gradient-based algorithm, the upper bound, i.e., minimum distance to an adversarial example, is $2100.45$ based on the $L^2$ distance metric. \correction{That is, manipulations imposed on the flows exceeding this upper bound may be \emph{unsafe}}. Figure~\ref{fig:perturbations_FloorGymnastics} (3rd row) shows some of such unsafe perturbations on each optical flow, which result in the misclassification of the video into $\mathsf{Front Crawl}$ with confidence $97.04\%$. As for the lower bound, we observe that, after $1500$ iterations of the admissible A* algorithm, the lower bound reaches $146.61$. That is, manipulations within this $L^2$ norm ball are absolutely \emph{safe}. Some of such safe perturbations can be found in the bottom row of Figure~\ref{fig:perturbations_FloorGymnastics}.

\end{document}